\documentclass[11pt]{article}

\PassOptionsToPackage{dvipsnames}{xcolor}

\usepackage[final]{acl}

\usepackage{times}
\usepackage{latexsym}

\usepackage[T1]{fontenc}
\usepackage[utf8]{inputenc}

\usepackage{microtype}

\usepackage{inconsolata}

\usepackage{graphicx}
\usepackage{hyperref}
\usepackage{url}

\usepackage{enumitem}
\usepackage{amsmath}
\usepackage[utf8]{inputenc} % allow utf-8 input
\usepackage[T1]{fontenc}    % use 8-bit T1 fonts
\usepackage{hyperref}       % hyperlinks
\usepackage{url}            % simple URL typesetting
\usepackage{booktabs}       % professional-quality tables
\usepackage{amsfonts}       % blackboard math symbols
\usepackage{amssymb}        % \varnothing and other AMS symbols
\usepackage{nicefrac}       % compact symbols for 1/2, etc.
\usepackage{microtype}      % microtypography
\usepackage{booktabs}
\usepackage{graphicx}
\usepackage{multirow}
\usepackage{subcaption}
\usepackage{listings}
\usepackage{lscape}
\usepackage{longtable}
\usepackage{placeins}
\usepackage{siunitx}
\usepackage{tcolorbox}
\usepackage{enumitem}
\usepackage{float}
\usepackage{xspace}
\usepackage{tabularx}
\usepackage{booktabs}

\usepackage{booktabs}
\usepackage{multirow}
\usepackage{makecell}
\usepackage{tabularx}

\tcbuselibrary{breakable}
\usepackage{colortbl}  % 用于背景色
\definecolor{mygreen}{HTML}{1E8449} % 用于正向提升的深绿色
\definecolor{myred}{HTML}{C0392B}
\definecolor{mygray}{gray}{0.92} % 极淡的灰色用于高亮行

\renewcommand{\arraystretch}{0.8}

\newtcolorbox{jsonexample}{
    colback=gray!10, % 轻微的背景色
    colframe=gray!50,
    boxsep=4pt, left=8pt, right=4pt, top=4pt, bottom=4pt,
    sharp corners,
    breakable, % 允许盒子内容跨页
    before upper={\ttfamily}
}

\newtcolorbox{rqbox}{
    colback=gray!5,      % 极淡的灰色背景
    colframe=gray!60,    % 边框颜色
    leftrule=3pt,        % 左边框加粗，制造“引用”感
    rightrule=0pt, toprule=0pt, bottomrule=0pt, % 其他边框隐藏
    boxsep=2pt, left=10pt, right=10pt, top=2pt, bottom=2pt,
    sharp corners,
    breakable,
    fontupper=\itshape   % 内容使用斜体，表示强调
}

\DeclareRobustCommand{\method}{\textsc{AdaptRubric\xspace}}

\title{Task-Adaptive Rubrics for GUI Reward Modeling}
\author{
 \textbf{Tao Xiong\textsuperscript{1}},
 \textbf{Xavier Hu\textsuperscript{1}},
 \textbf{Wenkai Wang\textsuperscript{1}},
 \textbf{Qinzhuo Wu\textsuperscript{2}},
\\
 \textbf{Changqiao Wu\textsuperscript{2}},
 \textbf{Pengzhi Gao\textsuperscript{2}},
 \textbf{Wei Liu\textsuperscript{2}},
 \textbf{Jian Luan \textsuperscript{2}},
 \textbf{Shengyu Zhang\textsuperscript{1,\ddag}}
\\
\\
 \textsuperscript{1}Zhejiang University,
 \textsuperscript{2}MiLM Plus, Xiaomi Inc.
\\
\\
\textbf{Correspondence:} \{xiongtao@zju.edu.cn,sy\_zhang@zju.edu.cn\}
}

\begin{document}
\maketitle

\renewcommand{\thefootnote}{}
\footnotetext{$^{\ddag}$Corresponding Author} 
\renewcommand{\thefootnote}{\arabic{footnote}}

\begin{abstract}

Recent studies on GUI agents have increasingly focused on \emph{outcome reward modeling}, which assigns outcome rewards by judging whether an executed trajectory satisfies the success criteria implied by the user instruction.
Existing GUI reward verifiers, however, often under-specify how these criteria should be constructed for \textbf{each task instance}.
Whether using generic rubric structures or implicit model reasoning, their judging criteria are not sufficiently task-adaptive: they can transfer checks across tasks, overlook concrete constraints in the current instruction, or become overly strict by enforcing unstated requirements.
To address this limitation, we propose \method, a Coarse-to-Fine Rubrics Framework that constructs task-adaptive judging criteria through a category-level coarse stage and an instance-level fine stage.
\method~performs \textbf{category-level coarse rubric retrieval} by routing the instruction to a GUI task family and retrieving reusable task-family criteria, then conducts \textbf{instance-level fine rubric generation} to surface compact cues for concrete values, scopes, and constraints in the current instruction.
Across offline reward evaluation and online reinforcement learning optimization, \method~consistently outperforms prior reward agents, improving F1 by 3.6 points over the baseline average under a matched image budget and yielding a 4.23-point task-success gain.
\end{abstract}

\section{Introduction}

Graphical User Interface (GUI) agents \citep{hu2025osagentssurveymllmbased, zhang2025instructiontuninglargelanguage, liu2025infiguiagentmultimodalgeneralistgui, wang2024gui} have emerged as a promising paradigm for automating realistic digital devices. Enhancing GUI agents relies heavily on reliable outcome reward models (ORMs), which can filter high-value trajectories \citep{xia2025agentrmenhancingagentgeneralization, chen2025scalingautonomousagentsautomatic, xiong2025guipraprocessrewardagent} from costly interactions, support benchmark evaluation and inference-time trajectory selection \citep{rawles2025androidworlddynamicbenchmarkingenvironment, xie2024osworldbenchmarkingmultimodalagents}, as well as provide reward signals \citep{xu2025mobilerlonlineagenticreinforcement, xu2026mobileagentv35multiplatformfundamentalgui, li2026osthemisscalablecriticframework} for reinforcement learning.

%Given the rule-based outcome verification is difficult to scale to open-ended settings. 
\begin{figure}[!t]
    \centering
    \includegraphics[width=1.0\linewidth]{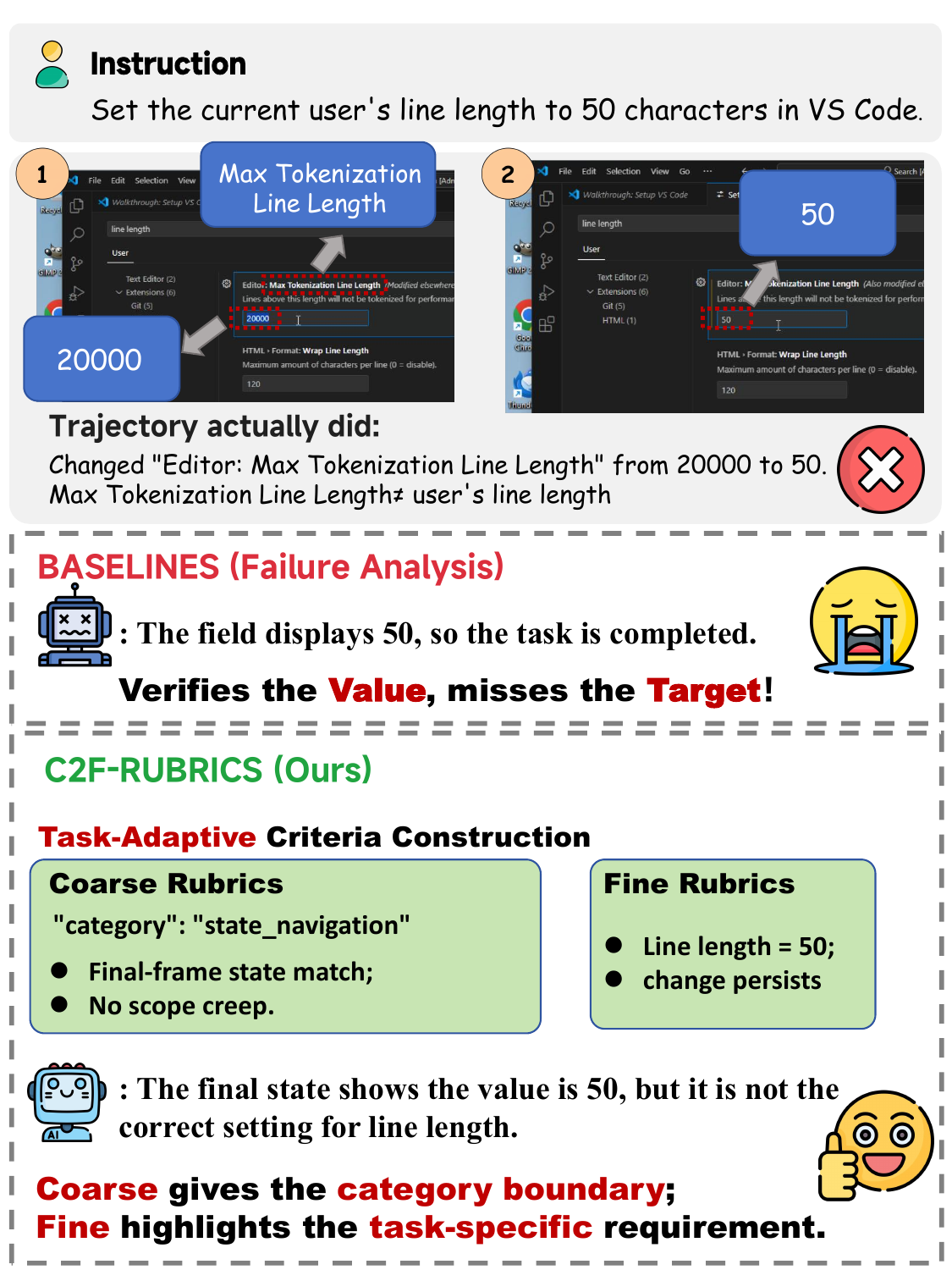}
    \caption{
    A motivating example for task-adaptive verification.
    The trajectory sets a related VS Code option to 50, but misses the requested line-length setting.
    While baselines verify only the surface value, \method~constructs coarse-to-fine criteria and correctly identifies the failure.
    }
    \label{fig:task}
\end{figure}

\begin{figure*}[!t]
    \centering
    \includegraphics[width=\linewidth]{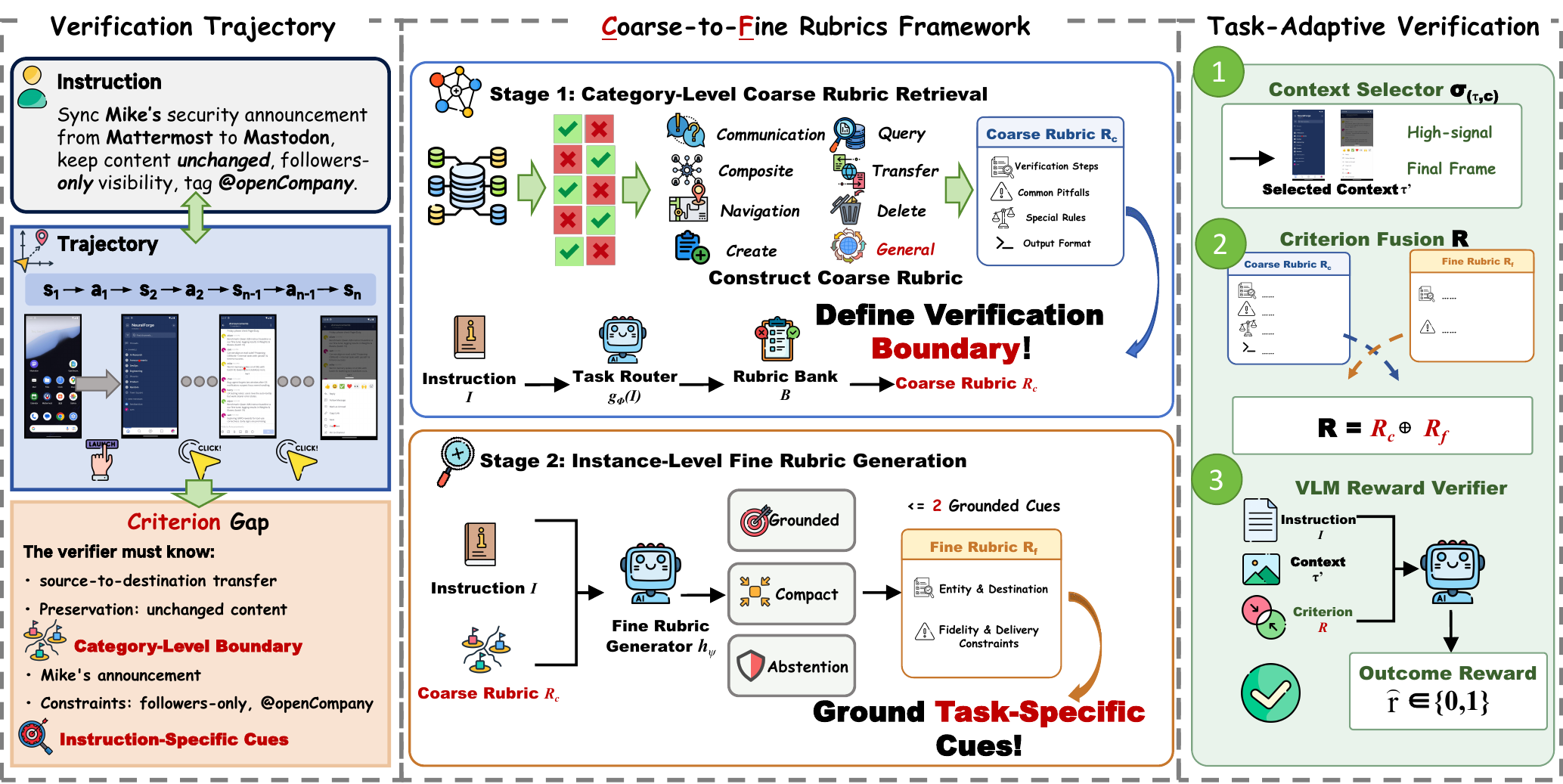}
    \caption{
    Overview of \method. The left panel motivates task-adaptive verification with a concrete trajectory example and the resulting criterion gap.
    The middle panel illustrates \textbf{coarse-to-fine} rubric construction, where \method\ retrieves a category-level coarse rubric from the rubric bank and augments it with compact instruction-specific cues.
    The right panel shows the final verification stage, where the fused criterion and selected trajectory context are passed to a VLM verifier for outcome reward prediction.
    }
    \label{fig:overview}
\end{figure*}

% To determine whether a GUI trajectory succeeds, a reward verifier must first identify the success criteria implied by the instruction. These criteria specify the goal to be achieved, the instance-specific constraints to satisfy, and the visual evidence needed to support the final judgment.
% However, existing GUI reward verifiers often lack clear and task-adaptive criterion construction. Static-template methods encode the criteria in a fixed rubric structure, which makes verification stable but keeps the rubric generic. Such a rubric may ask whether the task is completed, but it does not adapt its checks to the current instruction's target object, required value, operation scope, or output format. Other methods leave the criteria to the model's implicit reasoning during verification. This gives the verifier more flexibility, but without clear verification boundaries for different task categories, the model may overlook required instruction details or introduce constraints beyond the intended scope of the task. In both cases, the resulting criteria are not sufficiently task-adaptive.
To determine whether a GUI trajectory succeeds, a reward verifier must first identify the success criteria \citep{gupta2025carmodynamiccriteriageneration, xie2026autorubriclearningimplicitweights, gunjal2025rubricsrewardsreinforcementlearning} implied by the instruction. These criteria specify the goal to be achieved, the instance-specific constraints to satisfy, and the visual evidence needed to support the final judgment.
Existing GUI reward verifiers construct or obtain these criteria in two common ways. Static-template methods \citep{yang2025zeroguiautomatingonlinegui, qi2025webrltrainingllmweb, lai2025androidgenbuildingandroidlanguage, wang2025distrlasynchronousdistributedreinforcement, pan2024autonomousevaluationrefinementdigital} encode the criteria in a fixed rubric structure, which makes verification stable but keeps the rubric generic. Such a rubric may ask whether the task is completed, but it does not adapt its checks to the current instruction's target object, required value, operation scope, or output format. Other methods \citep{li2026osthemisscalablecriticframework, dai2026proreproactiverewardgui, cui2026agenticrewardmodelingverifying} leave the criteria to the model's implicit reasoning during verification. This gives the verifier more flexibility, but without clear verification boundaries for different task categories, the model may overlook required instruction details or introduce constraints beyond the intended scope of the task.

These limitations suggest that GUI reward verification requires task-adaptive criterion construction. The criteria should preserve verification boundaries for different task categories while adapting to the concrete requirements of each instruction instance. This allows the verifier to assess the trajectory against the task’s actual requirements, rather than a generic template or a loosely defined set of implicit checks.

We propose \method, a framework for task-adaptive criterion construction in GUI outcome reward modeling. \method~constructs explicit judging criteria through a category-level coarse stage and an instance-level fine stage. 
The coarse stage first routes each instruction to a GUI task category and retrieves a reusable rubric from a category-level rubric bank, where the rubric specifies verification steps, common pitfalls, and special rules for that category. 
% The coarse stage first routes each instruction to a GUI task category and retrieves a reusable rubric that specifies verification steps, common pitfalls, and special rules.
% The fine stage then generates a compact instance-level rubric grounded in the current instruction, turning concrete requirements into task-specific checking items. 
The fine stage then uses the current instruction and the retrieved rubric to generate compact instance-level rubric cues, turning concrete requirements into task-specific checking items. 
Finally, \method~fuses both components into a single criterion supplied to the VLM verifier, so the reward model preserves category-level verification boundaries while adapting to the current instruction.

We evaluate \method~on both offline reward discrimination and online reinforcement learning. On the offline GUI reward benchmark, \method~achieves 86.7\% accuracy and 86.6 F1, improving F1 by 3.6 points over the baseline average under a matched image budget. In online reinforcement learning experiments, using \method~as the reward verifier yields a 4.23-point absolute gain in task success rate and achieves the best result among compared reward agents. These results show that task-adaptive criterion construction improves trajectory-level reward judgment and transfers to downstream GUI agent optimization. Our contributions are summarized as follows:
\begin{itemize}[leftmargin=*,itemsep=1pt]
    \item We formulate task-adaptive criterion construction for GUI outcome reward modeling, where a verifier must derive explicit judging criteria from the user instruction before assigning reward.
    \item We introduce \method, a coarse-to-fine rubric framework that combines category-level rubrics with compact instance-level rubrics into a single task-adaptive criterion for VLM-based reward verification.
    % \item We evaluate \method~on offline reward discrimination and online reinforcement learning, showing improvements in both trajectory-level reward judgment and downstream GUI agent optimization.
    \item We evaluate \method~on offline reward discrimination and online reinforcement learning, achieving 86.7\% accuracy and 86.6 F1 on the offline benchmark and a 4.23-point absolute gain in online task success rate.
\end{itemize}

\section{Related Work}

\paragraph{GUI Agents for Task Automation.}
The rapid progress of multimodal large language models \citep{singh2026openaigpt5card, anthropic2025claude4, qwen35blog} has fundamentally transformed the landscape of autonomous GUI agents.
Early GUI agents commonly relied on structured interface representations, such as DOM/HTML trees \citep{gur2018learningnavigateweb,deng2023mind2webgeneralistagentweb} for web  pages and accessibility metadata for web or mobile interfaces~\citep{li2024effectsdatascaleui}. 
With the emergence of large multimodal language models, later work increasingly shifted toward screenshot-based observations, often augmented with Set-of-Mark-style \citep{yang2023setofmarkpromptingunleashesextraordinary} visual annotations to expose clickable regions for visual grounding.
To push their capability further, recent work post-trains these agents with supervised fine-tuning \citep{liu2025infiguir1advancingmultimodalgui, hong2024cogagentvisuallanguagemodel} and offline reinforcement learning~\citep{liu2025infiguir1advancingmultimodalgui, liu2024autoglmautonomousfoundationagents}.
However, offline data limits how much an agent can explore beyond the trajectories it has already seen.
Online reinforcement learning lets the agent interact with real environments, collect more diverse trajectories, and improve from a reward signal.
To provide scalable training data and a reliable reward for both settings, outcome reward modeling is gaining increasing attention.

\paragraph{Outcome Reward Modeling for GUI Agents.}
Reliable outcome reward modeling is critical for GUI agents.
A common approach is to use programmatic or rule-based evaluators \citep{rawles2025androidworlddynamicbenchmarkingenvironment, xie2024osworldbenchmarkingmultimodalagents, chen2025stepsuccessrateawaretrajectoryefficientpolicy} when the task can be deterministically checked, as in environments with executable assertions or handcrafted reward functions.
Such evaluators are precise but costly to design and hard to scale to open-ended GUI tasks.
Recent work therefore turns to model-based evaluators that estimate task success from screenshots, action histories, final states, or UI metadata.
One line of work directly applies LLM-as-a-judge, feeding the trajectory and the instruction into a single model to obtain a binary or graded reward~\citep{yang2025zeroguiautomatingonlinegui, qi2025webrltrainingllmweb, lai2025androidgenbuildingandroidlanguage, wang2025distrlasynchronousdistributedreinforcement, pan2024autonomousevaluationrefinementdigital}.
Another line improves verification reliability by decomposing the trajectory into milestones, adding verification steps, or introducing process-level rewards~\citep{li2026osthemisscalablecriticframework,zheng2026adaptivemilestonerewardgui, dai2026proreproactiverewardgui, cui2026agenticrewardmodelingverifying}.
These methods mainly change how evidence is collected, decomposed, or aggregated during verification.
We focus on a complementary issue: before the verifier evaluates any evidence, it needs a task-adaptive judging criterion derived from the instruction.
\method~addresses this by combining category-level rubrics with instance-level rubrics into an explicit criterion for GUI outcome reward verification.

\section{Preliminary}
\label{sec:preliminary}

A GUI agent interacts with an environment through a sequence of visual states and actions. Given a user instruction $\mathcal{I}$ and an initial screenshot $s_1$, the agent executes an action $a_t$ at each step and receives the next screenshot $s_{t+1}$ until termination. We denote the completed trajectory as
\begin{equation}
    \tau = (s_1,a_1,s_2,a_2,\ldots,s_T,a_T,s_{T+1}),
\end{equation}
where $s_t \in \mathcal{S}$ is a GUI screenshot, $a_t \in \mathcal{A}$ is the executed action, and $s_{T+1}$ is the terminal screenshot after the final action.

An outcome reward model (ORM) maps the instruction and trajectory to a binary success judgment,
\begin{equation}
    \hat{r} = \mathcal{M}(\mathcal{I}, \tau) \in \{0, 1\}.
    \label{eq:orm}
\end{equation}
% In this paper, the reward model is instantiated as a VLM-based verifier whose judgment depends on an explicit judging criterion. We denote this criterion by $\mathcal{R}$ and write
% \begin{equation}
%     \hat{r} = f_\theta(\mathcal{I}, \tau', \mathcal{R}) \in \{0,1\},
%     \label{eq:criterion_verifier}
% \end{equation}
% where $f_\theta$ is the VLM verifier, $\tau' \subseteq \tau$ is the selected trajectory context, and $\mathcal{R}$ specifies the success criteria used to judge the trajectory. The goal of task-adaptive criterion construction is to derive $\mathcal{R}$ for the current instruction before assigning reward.
Task-adaptive outcome reward modeling requires an explicit judging criterion that specifies the success conditions for the current instruction. We denote this criterion by $\mathcal{R}$ and write
\begin{equation}
    \hat{r} = f_\theta(\mathcal{I}, \tau', \mathcal{R}) \in \{0,1\},
    \label{eq:criterion_verifier}
\end{equation}
where $f_\theta$ is the verifier, $\tau' \subseteq \tau$ is the selected trajectory context, and $\mathcal{R}$ specifies the success criteria used to judge the trajectory. The goal of task-adaptive criterion construction is to derive $\mathcal{R}$ for the current instruction before assigning reward.

\section{Method}
\label{sec:method}

Figure~\ref{fig:overview} gives a compact overview of \method. As shown in the figure, Section~\ref{sec:method_coarse} introduces category-level coarse rubric retrieval, Section~\ref{sec:method_fine} describes instance-level fine rubric generation, and Section~\ref{sec:method_verification} presents criterion fusion and reward verification.

\subsection{Category-Level Coarse Rubric Retrieval}
\label{sec:method_coarse}

We construct the category-level rubric bank once before evaluation through an LLM-assisted induction procedure. Starting from a development trajectory pool $\mathcal{D}_{\mathrm{dev}}$ collected from existing GUI agent benchmarks, we sample successful and failed trajectories and ask an LLM to summarize the verification dimensions that separate successful completions from failures. We then group these dimensions into broad GUI task categories. Each category rubric is refined on held-out trajectories, and whenever the rubric judgment disagrees with the trajectory label, we ask the LLM to revise it.

The resulting taxonomy $\mathcal{C}$ contains $K{=}8$ categories, namely \texttt{info\_query}, \texttt{create\_modify}, \texttt{delete\_cleanup}, \texttt{communication}, \texttt{transfer}, \texttt{state\_navigation}, \texttt{composite\_workflow}, and \texttt{general}. The rubric bank is a fixed collection of category-level entries,
\begin{equation}
    \mathcal{B} = \{E_c=(m_c,S_c)\}_{c \in \mathcal{C}}.
\end{equation}
Each entry $E_c$ contains metadata $m_c$ and a set of natural-language rubric sections $S_c$. The metadata records the category label and prompt-assembly information, while the rubric sections specify verification steps, common pitfalls, special rules, and output format:
\begin{equation}
    S_c = \{S_c^{\mathrm{step}}, S_c^{\mathrm{pitfall}}, S_c^{\mathrm{rule}}, S_c^{\mathrm{format}}\},
\end{equation}
During verification, \method~renders $S_c$ into the category-level coarse rubric $R_c$. The taxonomy and rubric dimensions are summarized in Appendix~\ref{app:taxonomy}.
At inference time, a task router predicts the GUI task category from the instruction,
\begin{equation}
    c = g_{\phi}(\mathcal{I}), \qquad c \in \mathcal{C}.
    \label{eq:router}
\end{equation}
\method~then retrieves and renders the corresponding bank entry,
\begin{equation}
    \begin{gathered}
        E_c = \operatorname{Lookup}(\mathcal{B}, c), \quad
        S_c = \operatorname{Sections}(E_c), \\
        R_c = \operatorname{Render}(S_c).
    \end{gathered}
    \label{eq:coarse_lookup}
\end{equation}
We use top-1 lookup and if the router does not identify a specialized category, \method~falls back to the general entry $E_{\texttt{general}}$.

\subsection{Instance-Level Fine Rubric Generation}
\label{sec:method_fine}

The fine stage constructs an instance-level fine rubric for the current instruction. Given the instruction $\mathcal{I}$, the predicted category $c$, and the retrieved coarse rubric $R_c$, a rubric generator produces
\begin{equation}
    R_f = h_{\psi}(\mathcal{I}, c, R_c), \qquad |R_f| \le 2,
    \label{eq:fine}
\end{equation}
where $R_f$ is a compact list of fine-grained rubric items. The generator receives the coarse rubric as context, but it is not asked to rewrite $R_c$ or fill its sections. Instead, it generates only additional instance-level checks that are grounded in the current instruction.
% The generation process follows three constraints. First, each fine rubric item must be supported by an explicit phrase, value, or constraint in the user instruction. Second, $R_f$ contains at most two items, so the fine rubric highlights only the most useful instance-level requirements rather than becoming another full checklist. Third, the generator may abstain and return $R_f=\varnothing$ when no reliable fine rubric item is needed. These constraints keep the fine rubric compact and reduce the chance of adding requirements that are not specified by the instruction.
The generator follows three constraints:
\begin{itemize}[leftmargin=*,itemsep=1pt]
    \item \textbf{Instruction grounding.} Each fine rubric item must be supported by an explicit phrase, value, or constraint in the user instruction.
    \item \textbf{Compactness.} $R_f$ contains at most two items, so the fine rubric highlights only the most useful instance-level requirements rather than becoming another full checklist.
    \item \textbf{Abstention.} The generator may return $R_f=\varnothing$ when no reliable fine rubric item is needed.
\end{itemize}
These constraints keep the fine rubric compact and reduce the chance of adding requirements that are not specified by the instruction.

\newcolumntype{N}{>{\centering\arraybackslash}p{2.25em}} % 数值列等宽
\newcommand{\gsep}{\hskip 4pt}
\begin{table*}[t]
  \centering
  \scriptsize
  \setlength{\tabcolsep}{1.8pt}
  \renewcommand{\arraystretch}{1.1}

  \resizebox{\textwidth}{!}{%
  \begin{tabular}{l
      N N @{\gsep}
      N N @{\gsep}
      N N @{\gsep}
      N N @{\gsep}
      N N @{\gsep}
      N N N N
    }
    \toprule
    \multirow{2}{*}{\textbf{Model}} &
    \multicolumn{2}{c}{\textbf{Ubuntu}} &
    \multicolumn{2}{c}{\textbf{Mobile}} &
    \multicolumn{2}{c}{\textbf{Windows}} &
    \multicolumn{2}{c}{\textbf{macOS}} &
    \multicolumn{2}{c}{\textbf{Web}} &
    \multicolumn{4}{c}{\textbf{Overall}} \\
    \cmidrule(lr){2-3}\cmidrule(lr){4-5}\cmidrule(lr){6-7}\cmidrule(lr){8-9}\cmidrule(lr){10-11}\cmidrule(lr){12-15}
    & \textbf{Acc} & \textbf{F1} &
      \textbf{Acc} & \textbf{F1} &
      \textbf{Acc} & \textbf{F1} &
      \textbf{Acc} & \textbf{F1} &
      \textbf{Acc} & \textbf{F1} &
      \textbf{Acc} & \textbf{Prec} & \textbf{Rec} & \textbf{F1} \\
    \midrule

    \rowcolor{gray!20} \multicolumn{15}{c}{\textbf{ZeroGUI}} \\
    Qwen3-VL-4B-Instruct & 83.8 & 84.0 & 74.5 & 76.2 & 80.3 & 75.6 & 90.9 & 78.8 & 81.1 & 83.2 & 82.0 & 83.2 & 80.0 & 81.6 \\
    Qwen3-VL-8B-Instruct & 84.6 & 85.0 & 83.0 & 84.5 & 79.3 & 74.4 & 94.8 & 88.2 & 78.4 & 80.6 & 83.3 & 84.1 & 81.9 & 83.0 \\
    Qwen3-VL-32B-Instruct & 84.6 & 85.2 & 83.5 & 85.0 & 80.3 & 75.6 & \underline{96.1} & 90.3 & 82.1 & 83.5 & 84.1 & 84.6 & 83.1 & 83.9 \\
    Qwen3-VL-235B-A22B-Instruct & 86.9 & 87.3 & 84.0 & 85.8 & 84.5 & 80.9 & \underline{96.1} & 90.3 & 87.4 & 88.6 & 86.7 & 87.2 & 85.9 & 86.5 \\
    Qwen3.5-122B-A10B & 85.8 & 86.5 & 82.4 & 83.7 & 83.6 & 80.2 & \underline{96.1} & 90.9 & 80.0 & 82.7 & 84.8 & 84.1 & 85.6 & 84.8 \\
    Qwen3.6-27B & 86.2 & 87.5 & 80.9 & 83.5 & 83.6 & 81.5 & 94.8 & 88.9 & 82.1 & 84.7 & 85.0 & 81.2 & \underline{90.9} & 85.8 \\
    Gemini 3 Flash & \underline{88.5} & \underline{89.0} & 80.3 & 80.6 & \underline{87.8} & \underline{85.7} & \textbf{97.4} & \underline{93.8} & 87.9 & 88.3 & 87.7 & 89.2 & 85.7 & 87.4 \\
    Gemini 3.1 Flash-Lite & 87.2 & 87.9 & 80.3 & 81.6 & 85.4 & 82.3 & 94.8 & 87.5 & 85.3 & 86.8 & 86.2 & 85.9 & 86.3 & 86.1 \\
    \rowcolor{gray!10} \textit{Mean} & 86.0 & 86.5 & 81.1 & 82.6 & 83.1 & 79.5 & 95.1 & 88.6 & 83.0 & 84.8 & 85.0 & 84.9 & 84.9 & 84.9 \\

    \midrule
    \rowcolor{gray!20} \multicolumn{15}{c}{\textbf{OS-Themis}} \\
    Qwen3-VL-4B-Instruct & 72.6 & 71.4 & 79.3 & 78.9 & 75.1 & 68.3 & 84.4 & 57.1 & 80.0 & 81.9 & 75.5 & 79.5 & 68.3 & 73.5 \\
    Qwen3-VL-8B-Instruct & 76.2 & 75.0 & 84.0 & 83.7 & 78.4 & 72.0 & 83.1 & 51.9 & 81.6 & 82.4 & 78.7 & 84.6 & 69.9 & 76.5 \\
    Qwen3-VL-32B-Instruct & 77.1 & 74.6 & 81.9 & 80.7 & 76.5 & 65.3 & 90.9 & 77.4 & 83.7 & 81.9 & 79.3 & 91.6 & 64.1 & 75.5 \\
    Qwen3-VL-235B-A22B-Instruct & 86.4 & 86.8 & \textbf{93.6} & \underline{93.7} & 77.5 & 69.6 & 93.5 & 82.8 & 91.6 & 91.9 & 87.1 & 90.5 & 82.7 & 86.4 \\
    Qwen3.5-122B-A10B & 85.8 & 85.8 & 88.3 & 88.2 & 79.8 & 73.0 & 88.3 & 66.7 & 79.0 & 75.6 & 84.5 & \underline{92.0} & 75.3 & 82.8 \\
    Qwen3.6-27B & 87.9 & 88.5 & \textbf{93.6} & \textbf{93.9} & \textbf{88.3} & \textbf{86.9} & \underline{96.1} & 90.3 & \underline{92.1} & 92.1 & \underline{89.7} & 90.2 & 89.0 & \underline{89.6} \\
    Gemini 3 Flash & 81.5 & 81.5 & 84.0 & 84.2 & 86.9 & 84.4 & 88.3 & 71.0 & 80.5 & 77.3 & 82.9 & 88.1 & 75.9 & 81.5 \\
    Gemini 3.1 Flash-Lite & 82.9 & 83.3 & 82.4 & 82.7 & 87.3 & 84.2 & 90.9 & 75.9 & 84.2 & 83.3 & 84.1 & 87.7 & 79.1 & 83.2 \\
    \rowcolor{gray!10} \textit{Mean} & 81.3 & 80.9 & 85.9 & 85.8 & 81.2 & 75.5 & 89.4 & 71.6 & 84.1 & 83.3 & 82.7 & 88.0 & 75.5 & 81.1 \\

    \midrule
    \rowcolor{gray!20} \multicolumn{15}{c}{\textbf{\method}} \\
    Qwen3-VL-4B-Instruct & 84.3 & 85.6 & 80.9 & 81.1 & 83.1 & 79.5 & \textbf{97.4} & \textbf{94.1} & 82.1 & 84.5 & 84.1 & 82.8 & 85.9 & 84.3 \\
    Qwen3-VL-8B-Instruct & 83.9 & 85.1 & 83.5 & 84.6 & 81.7 & 79.1 & \textbf{97.4} & \textbf{94.1} & 81.6 & 83.3 & 84.0 & 82.6 & 85.9 & 84.2 \\
    Qwen3-VL-32B-Instruct & 86.5 & 86.6 & 85.6 & 85.9 & 80.3 & 75.0 & 93.5 & 83.9 & 86.8 & 87.6 & 85.9 & 89.3 & 81.3 & 85.1 \\
    Qwen3-VL-235B-A22B-Instruct & 86.1 & 86.7 & 87.8 & 88.9 & 84.5 & 81.4 & 94.8 & 87.5 & 88.9 & 90.0 & 86.9 & 87.0 & 86.7 & 86.8 \\
    Qwen3.5-122B-A10B & 86.2 & 86.8 & 89.9 & 90.4 & 81.7 & 77.2 & 94.8 & 88.2 & 88.9 & 89.8 & 86.9 & 87.8 & 85.4 & 86.6 \\
    Qwen3.6-27B & 87.3 & 88.5 & \underline{91.0} & 91.6 & 84.5 & 82.5 & 93.5 & 85.7 & 91.6 & \underline{92.4} & 88.3 & 85.4 & \textbf{92.1} & 88.7 \\
    Gemini 3 Flash & \textbf{90.4} & \textbf{90.7} & \underline{91.0} & 91.4 & 87.3 & 84.7 & 94.8 & 88.9 & \textbf{94.7} & \textbf{95.0} & \textbf{90.8} & \textbf{92.3} & 89.0 & \textbf{90.6} \\
    Gemini 3.1 Flash-Lite & 86.4 & 87.1 & 83.0 & 83.3 & 85.0 & 81.6 & 94.8 & 87.5 & 88.9 & 89.7 & 86.5 & 87.2 & 85.4 & 86.3 \\
    \rowcolor{gray!10} \textit{Mean} & 86.4 & 87.1 & 86.6 & 87.2 & 83.5 & 80.1 & 95.1 & 88.7 & 87.9 & 89.0 & 86.7 & 86.8 & 86.5 & 86.6 \\
    \bottomrule
  \end{tabular}%
  }
  \caption{Offline reward discrimination results on OGRBench. Each method reports Accuracy (Acc) and F1-score (F1) for each platform; Overall columns include Acc, Precision (Prec), Recall (Rec), and F1. ZeroGUI uses the last-10 image budget. In each column, \textbf{bold} marks the best value across all $3{\times}8$ method-backbone cells, and \underline{underline} marks the second-best (computed separately for the \textit{Mean} rows).}
  \label{tab:OmniGUIRewardBench_Simple}
\end{table*}

\subsection{Criterion Fusion and Reward Verification}
\label{sec:method_verification}

After obtaining the category-level coarse rubric and the instance-level fine rubric, \method~assembles the final judging criterion as
\begin{equation}
    \mathcal{R} = \Phi(R_c, R_f) = R_c \oplus R_f.
    \label{eq:fuse}
\end{equation}
The operator $\oplus$ keeps $R_c$ as the main body of the criterion and appends $R_f$ as a separated task-specific block. If the fine stage abstains, the final criterion reduces to the coarse rubric, $\Phi(R_c,\varnothing)=R_c$. The fused criterion therefore preserves the category-level verification boundary while adding the instruction-specific fine rubric when available.

Before verification, \method~selects a compact trajectory context from the full GUI trajectory,
\begin{equation}
    \tau'=\sigma(\tau,c),\qquad |\tau'|\le M.
    \label{eq:select}
\end{equation}
The selector keeps screenshot-bearing steps that are useful for outcome judgment, including the initial and final states and frames around high-signal actions such as text entry, submission, or state-changing operations. The selected context, the instruction, and the fused criterion are then passed to the verifier,
\begin{equation}
    \hat{r} = f_\theta(\mathcal{I}, \tau', \mathcal{R}) \in \{0,1\}.
    \label{eq:verifier}
\end{equation}
The verifier outputs a binary reward in a parsable format. Thus, \method~changes the judging criterion and trajectory context supplied to the verifier, while keeping the verifier architecture unchanged.

\section{Experiment}

\subsection{Offline Evaluation}
We evaluate offline GUI reward discrimination on OmniGUIRewardBench (OGRBench) \citep{li2026osthemisscalablecriticframework}, a cross-platform benchmark for testing whether a reward verifier can correctly judge trajectory-level task success.

\begin{figure*}[!t]
    \centering
    \begin{subfigure}[t]{0.48\textwidth}
        \centering
        \includegraphics[width=\linewidth]{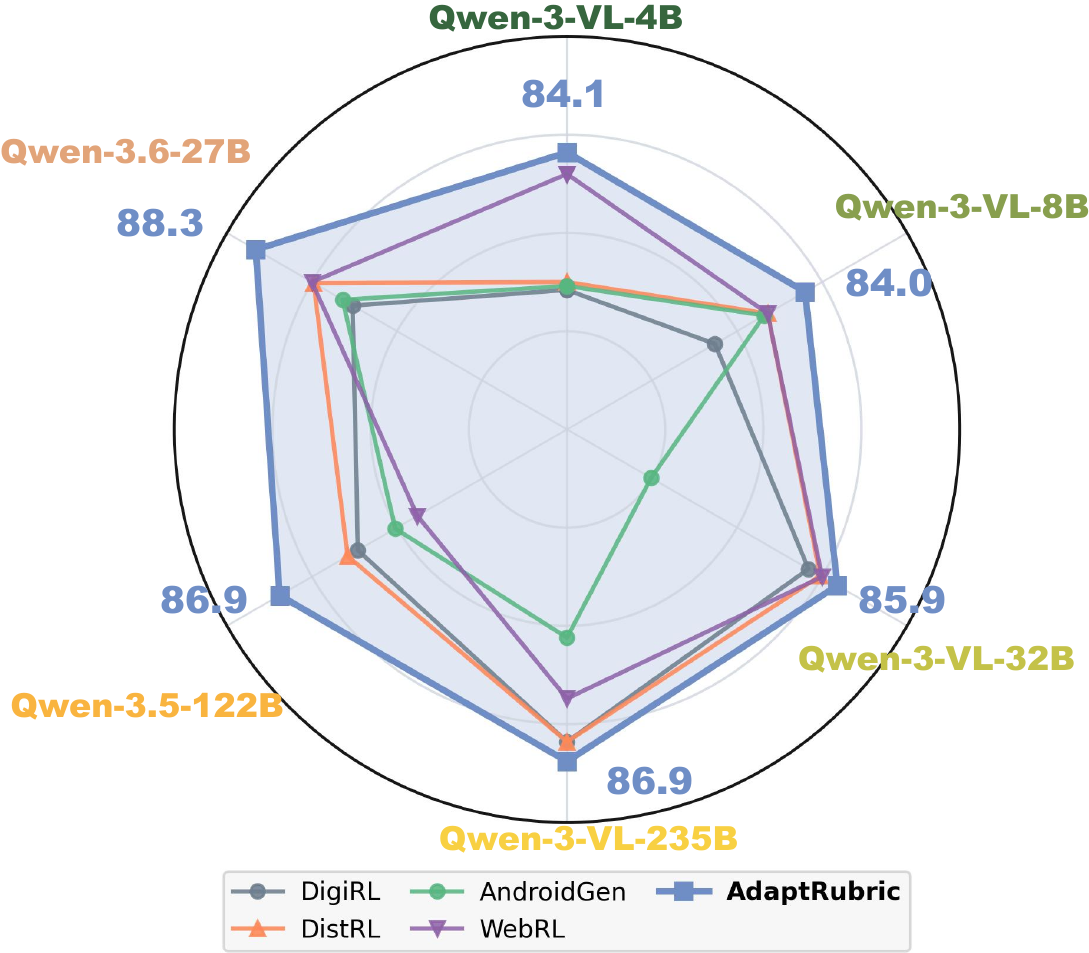}
        \caption{Overall Accuracy.}
        \label{fig:offline_radar_acc}
    \end{subfigure}
    \hfill
    \begin{subfigure}[t]{0.48\textwidth}
        \centering
        \includegraphics[width=\linewidth]{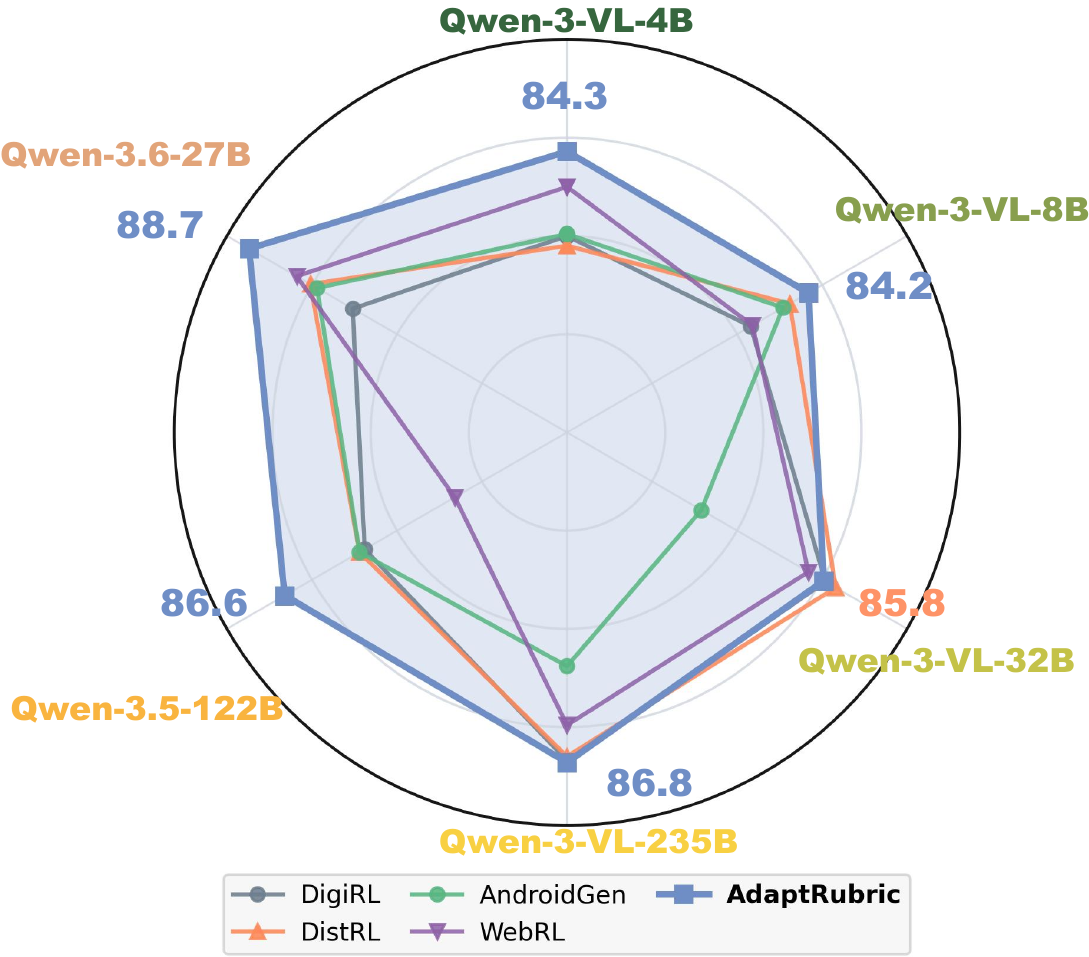}
        \caption{Overall F1.}
        \label{fig:offline_radar_f1}
    \end{subfigure}
    \caption{
    Cross-backbone radar comparison on OGRBench under the matched ten-screenshot budget.
    Each axis corresponds to one Qwen-family judge backbone, and each curve corresponds to one reward verifier.
    }
    \label{fig:offline_radar}
\end{figure*}

\textbf{Benchmark.}
OGRBench contains 1,409 trajectories from five environments: OSWorld \citep{xie2024osworldbenchmarkingmultimodalagents} for Ubuntu, AndroidWorld \citep{rawles2025androidworlddynamicbenchmarkingenvironment} for mobile Android,  \citep{bonatti2024windowsagentarenaevaluating} for Windows, macOSArena \citep{wang2025mmbenchgui} for macOS, and WebArena-Lite-v2 \citep{wang2025mmbenchgui, zhou2023webarena} for web tasks.
The benchmark is nearly balanced overall, with 700 positive and 709 negative trajectories. % Each example contains a complete GUI interaction trajectory with screenshots and agent outputs, paired with a binary success label from the corresponding benchmark evaluator.

\textbf{Baselines.}
We compare with six representative GUI reward verification frameworks: DigiRL~\citep{bai2024digirltraininginthewilddevicecontrol}, DistRL~\citep{wang2025distrlasynchronousdistributedreinforcement}, AndroidGen~\citep{lai2025androidgenbuildingandroidlanguage}, WebRL~\citep{qi2025webrltrainingllmweb}, ZeroGUI~\citep{yang2025zeroguiautomatingonlinegui}, and OS-Themis~\citep{li2026osthemisscalablecriticframework}.
To align their trajectory inputs with \method, all offline reward verifiers are evaluated under the same ten-screenshot budget.
Appendix~\ref{app:additional_matched_budget} details each baseline's original trajectory-context setting and our unified matched-budget instantiation, and Appendix~\ref{app:image_budget_analysis} analyzes the effect of image budget.

For judge backbones, we evaluate open-source Qwen models from the Qwen3-VL \citep{Qwen3-VL}, Qwen3.5 \citep{qwen35blog}, and Qwen3.6 \citep{qwen3.6-27b} series, as well as closed-source Gemini models \citep{geminiteam2025geminifamilyhighlycapable}.

\textbf{Metrics.}
For offline reward discrimination, we report Accuracy, Precision, Recall, and F1.
% Precision reflects how often predicted-success trajectories are truly successful, which is important when reward labels are used for data filtering or reinforcement learning.
% Recall measures how many successful trajectories are recovered, and F1 summarizes the precision-recall trade-off.
% For online reinforcement learning and reward-guided trajectory selection, we report task success rate or success-rate gains.

% 完成该部分实验结果的分析
\textbf{Main Results.}
Table~\ref{tab:OmniGUIRewardBench_Simple} reports the main OGRBench comparison with ZeroGUI and OS-Themis across all eight judge backbones.
In this setting, \method~achieves the \textbf{best average} performance, with 86.7\% accuracy and 86.6 F1.
Compared with the average of the two baselines under the main protocol, \method~improves the four overall metrics by 3.3 points on average, including 2.9 points in accuracy and 3.6 points in F1.
The main gain comes from recall: the baseline average is 80.2\%, while \method~raises recall to 86.5\% with a comparable precision of 86.8\%.
This indicates that \textbf{task-adaptive criteria} help the verifier recover more successful trajectories while still controlling false positives.
The improvement is also \textbf{consistent across platforms}, where \method~achieves the best average F1 on Ubuntu, Android, Windows, macOS, and Web tasks.
These results support our core claim that combining category-level rubrics with instance-level fine rubrics yields more reliable trajectory-level reward judgment than generic terminal-state judging or implicit multi-step verification.

Figure~\ref{fig:offline_radar} further compares \method~with four additional baselines, DigiRL, DistRL, AndroidGen, and WebRL, using the open-source Qwen-family judge backbones.
This expanded comparison tests whether the same advantage holds against a broader set of reward-agent prompts under the matched ten-screenshot budget.
\method~achieves the best accuracy on all backbones and the best F1 on five of six backbones, showing that the gain is not tied to a single judge scale.
The detailed values in Appendix~\ref{app:additional_matched_budget} show that the expanded-baseline mean F1 of \method~remains higher than all four additional baselines.

\FloatBarrier

\subsection{Online Reinforcement Learning}
We further evaluate whether the offline reward advantage transfers to downstream GUI agent optimization.

\textbf{Setup.}
We conduct online RL training using the ClawGUI \citep{tang2026clawguiunifiedframeworktraining} framework on the MobileWorld \citep{kong2025mobileworldbenchmarkingautonomousmobile} environment.
The policy backbone is MAI-UI-8B \citep{zhou2025maiuitechnicalreportrealworld}, and the policy is optimized with  Group Relative Policy Optimization (GRPO) \citep{shao2024deepseekmathpushinglimitsmathematical}.
For each task, GRPO samples four rollouts as a group, and the reward agent assigns an outcome reward to each completed trajectory.
We keep the training protocol fixed and replace only the reward agent, comparing DigiRL, ZeroGUI, OS-Themis, and \method.
For \method, the reward verifier uses Qwen3-VL-8B-Instruct and observes at most ten trajectory screenshots.
All runs use a maximum episode length of 50 steps and train for 2 epochs; other hyperparameters are reported in Appendix~\ref{app:online_rl_details}.

\textbf{Main Results.}
Table~\ref{tab:maui8b_reward_agent} reports online reinforcement learning results when different reward agents provide training rewards for MAI-UI-8B.
Without an external reward agent, MAI-UI-8B reaches 19.70\% task success.
Using \method~as the reward verifier increases success to 23.93\%, a \textbf{4.23-point} absolute gain and the best result among compared reward agents.
This suggests that the criterion constructed by \method~is not only better at offline trajectory discrimination, but also provides a more useful reward signal for policy improvement.

\begin{table}[!b]
\centering
\scriptsize
\renewcommand{\arraystretch}{1.02}
\setlength{\tabcolsep}{3.5pt}
\resizebox{0.95\columnwidth}{!}{
\begin{tabular}{@{}llcc@{}}
\toprule
\textbf{Backbone} & \textbf{Reward Agent} & \textbf{SR (\%)} & \textbf{$\Delta$} \\
\midrule
\multirow{5}{*}{\textbf{MAI-UI-8B}}
& --        & 19.70$^\dagger$ & -- \\
& DigiRL    & 23.08 & \textcolor{mygreen}{$+3.38$} \\
& ZeroGUI   & 22.22 & \textcolor{mygreen}{$+2.52$} \\
& OS-Themis & 22.22 & \textcolor{mygreen}{$+2.52$} \\
\cmidrule(lr){2-4}
& \textbf{\method}
& $\mathbf{23.93 \pm 0.85}$
& \textcolor{mygreen}{$\mathbf{+4.23}$} \\
\bottomrule
\end{tabular}
}
\caption{Online reinforcement training results on MAI-UI-8B. We report the success rate (SR) after training with different reward agents. $^\dagger$ denotes the result from \citet{tang2026clawguiunifiedframeworktraining}.}
\label{tab:maui8b_reward_agent}
\end{table}

% \begin{figure}[!htbp]
%     \centering
%     \includegraphics[width=1.0\linewidth]{figures/fig_earlystop_v2.pdf}
%     \caption{EarlyStop SR@$N$ on heterogeneous AndroidWorld pool (113 tasks, $N{=}1{,}3{,}5{,}7$).}
%     \label{fig:earlystop}
% \end{figure}

% \begin{figure}[!htbp]
%     \centering
%     \includegraphics[width=1.0\linewidth]{figures/fig_bestofn_v2.pdf}
%     \caption{BestOfN SR@$N$ on heterogeneous AndroidWorld pool (113 tasks, $N{=}1{,}2{,}4{,}6{,}8$).}
%     \label{fig:bestofn}
% \end{figure}
\begin{figure*}[t]
    \centering
    \begin{subfigure}[t]{0.48\textwidth}
        \centering
        \includegraphics[width=\linewidth]{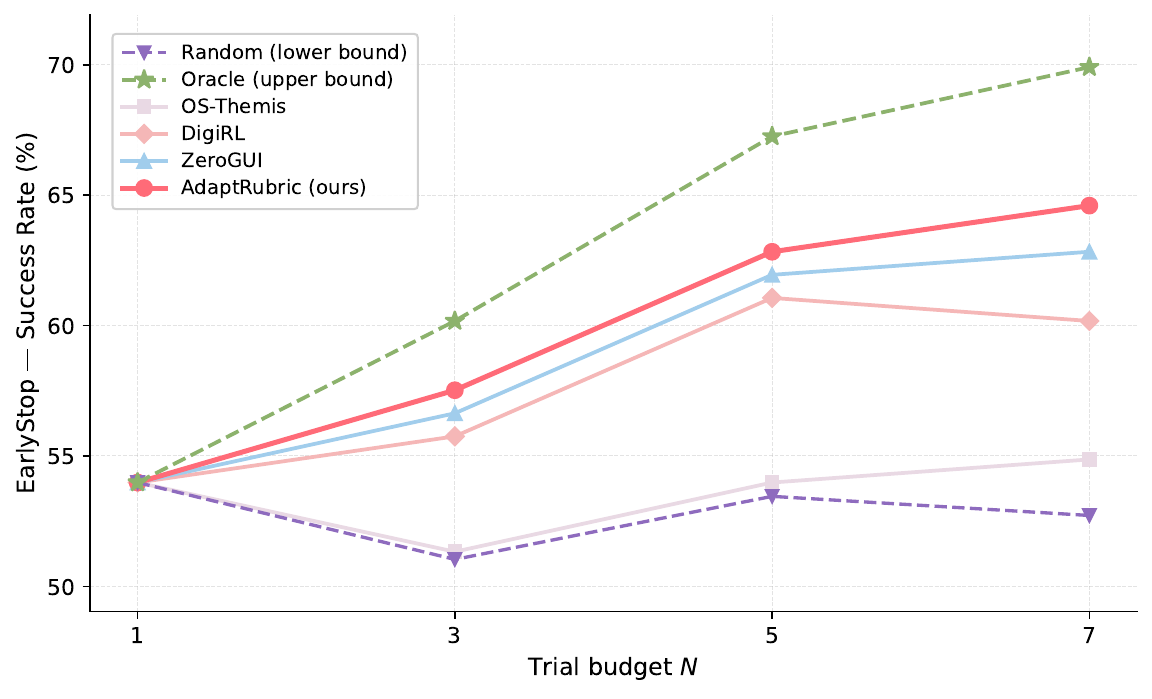}
        \caption{EarlyStop SR@$N$ ($N{=}1{,}3{,}5{,}7$).}
        \label{fig:earlystop}
    \end{subfigure}
    \hfill
    \begin{subfigure}[t]{0.48\textwidth}
        \centering
        \includegraphics[width=\linewidth]{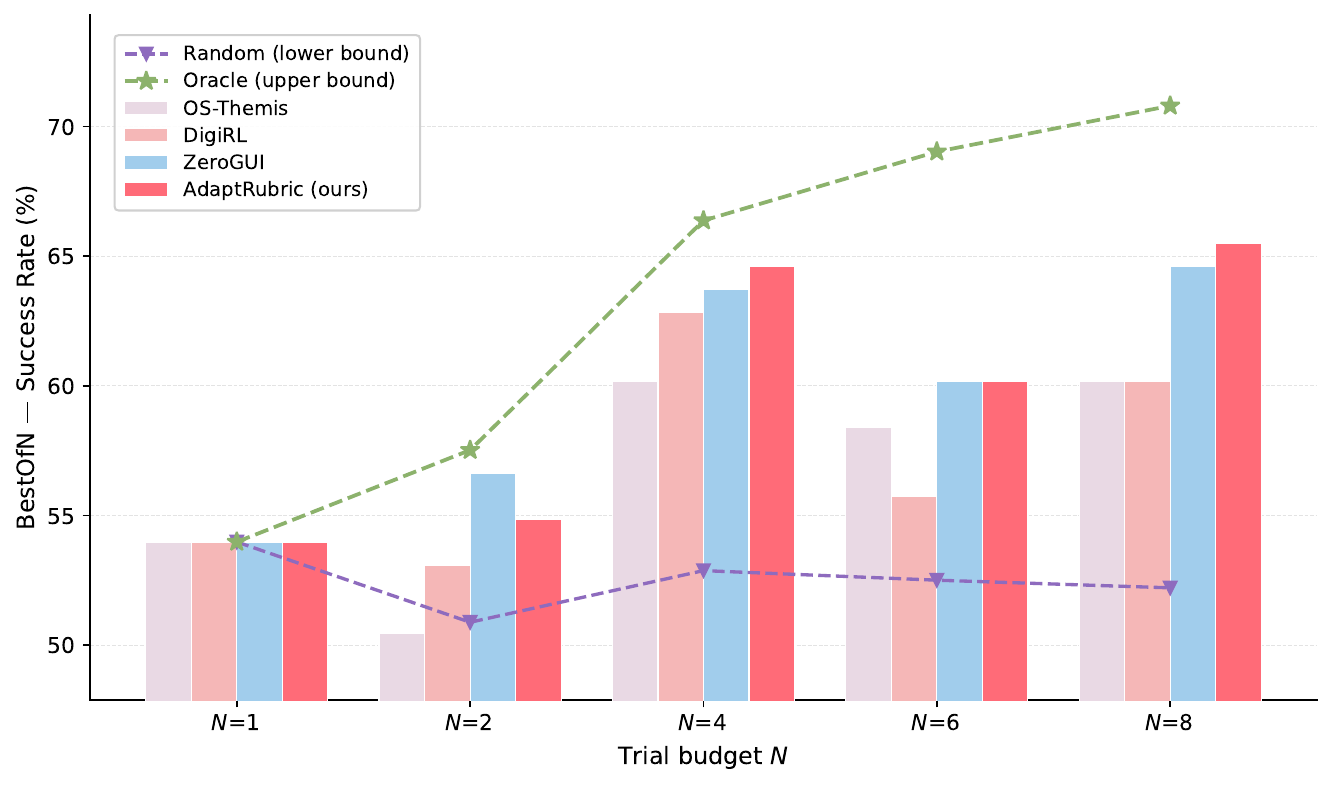}
        \caption{BestOfN SR@$N$ ($N{=}1{,}2{,}4{,}6{,}8$).}
        \label{fig:bestofn}
    \end{subfigure}

    \caption{Test-time scaling results on the heterogeneous AndroidWorld pool of 113 tasks.}
    \label{fig:test_time_scaling}
\end{figure*}

\begin{table}[t]
\centering
\small
\setlength{\tabcolsep}{3pt}
\renewcommand{\arraystretch}{1.1}
\resizebox{\columnwidth}{!}{%
\begin{tabular}{@{}lccccc@{}}
\toprule
\textbf{Method}
  & \textbf{ES@7}
  & \textbf{BoN@8}
  & \textbf{Acc.}
  & \textbf{F1}
  & \textbf{FPR} \\
\midrule
OS-Themis & +2.15           & +7.97            & 75.75          & 72.92          & \textbf{10.81} \\
DigiRL    & +7.46           & +7.97            & 80.80          & 81.90          & 22.71          \\
ZeroGUI   & +10.11          & +12.39           & 86.55          & 87.16          & 15.38          \\
\method   & \textbf{+11.88} & \textbf{+13.28}  & \textbf{88.14} & \textbf{88.41} & \underline{11.17} \\
\bottomrule
\end{tabular}%
}
\caption{
Heterogeneous-pool offline reward-guided test-time scaling.
ES@7 (EarlyStop) and BoN@8 (BestOfN) report success-rate gains over Random in percentage points.
Acc., F1, and FPR (false-positive rate) are computed over trajectory-level reward predictions.
}
\label{tab:offline_tts_results}
\end{table}

\subsection{Reward-Guided Test-Time Scaling}
We test whether reward verification helps select successful trajectories when multiple candidates are sampled for the same task.

\textbf{Setup.}
We evaluate on 113 AndroidWorld tasks, with ten candidate trajectories sampled for each task.
We use this offline candidate pool to simulate online test-time scaling, so that all verifiers select from the same trajectories and the dynamic execution environment remains controlled.
To create a discriminative candidate pool, we mix eight rollouts from Qwen3-VL-8B-Instruct and two rollouts from Qwen3-VL-235B-A22B-Instruct. Details are provided in Appendix~\ref{app:tts_details}.
All reward verifiers share Qwen3-VL-32B-Instruct as the judge and consume the same per-task shuffled trajectory stream.
We evaluate two selection protocols: \textbf{EarlyStop@N} scans the first $N$ trajectories sequentially and stops at the first one predicted as successful (online, low-latency setting), while \textbf{BestOfN@N} scores all $N$ candidates and returns the highest-scored one (offline, batch setting).
All verifiers in this comparison output binary scores, so multiple candidates often share the same score under BestOfN. In this case, we always pick the last tied candidate as the final selection.
Both gains are reported relative to Random selection.

\textbf{Overall results.}
Table~\ref{tab:offline_tts_results} summarizes performance under the two selection protocols.
\method~achieves the largest gains over Random, improving EarlyStop@7 by \textbf{$+11.88$} points and BestOfN@8 by \textbf{$+13.28$} points.
It also obtains the strongest trajectory-level reward prediction results, with \textbf{88.14} accuracy and \textbf{88.41} F1.
At the same time, its false-positive rate remains low at 11.17, close to the most conservative baseline OS-Themis.

\textbf{EarlyStop: low-latency selection.}
Figure~\ref{fig:earlystop} reports EarlyStop SR@$N$ as the per-task budget grows from 1 to 7.
\method~is the strongest method at every budget from $N{=}3$ onward.
At $N{=}7$, it reaches \textbf{$64.60\%$} task success, exceeding ZeroGUI by 1.77 points, DigiRL by 4.42 points, and OS-Themis by 9.73 points.
The Oracle upper bound at the same budget is \textbf{$69.91\%$}, so \method~closes about \textbf{$69\%$} of the gap between Random selection and the Oracle.
Because EarlyStop commits to the first accepted trajectory, it favors verifiers that can avoid false-positive acceptances.

\textbf{BestOfN: batch selection.}
Figure~\ref{fig:bestofn} reports BestOfN SR@$N$ for $N \in \{1,2,4,6,8\}$.
\method~achieves the best result at $N{=}4$ with \textbf{$64.60\%$} task success and at $N{=}8$ with \textbf{$65.49\%$} task success, and it ties ZeroGUI at $N{=}6$.
The only weaker point is $N{=}2$, where binary reward outputs often produce ties and the fixed tie-breaking rule has a large effect on the selected trajectory.
DigiRL decreases as the candidate set grows from $N{=}4$ to $N{=}6$, consistent with the difficulty of judging longer visual contexts.
Overall, the batch setting amplifies the advantage of \method, because reward errors accumulate when the verifier must compare many candidates.

% Local marks for this ablation table (from amssymb, already loaded).
\providecommand{\cmark}{\textcolor{mygreen}{\ding{51}}}
\providecommand{\xmark}{\textcolor{myred}{\ding{55}}}
% Fallback if pifont/ding is not available: use math symbols.
\providecommand{\yes}{\textcolor{mygreen}{$\checkmark$}}
\providecommand{\no}{\textcolor{myred}{$\times$}}

\begin{table}[t]
\centering
\small
\renewcommand{\arraystretch}{1.15}
\setlength{\tabcolsep}{4pt}
\resizebox{\columnwidth}{!}{%
\begin{tabular}{@{}cc l cc cc@{}}
\toprule
\textbf{Coarse} & \textbf{Fine} & \textbf{Setting}
  & \textbf{Acc.} & $\Delta$\textbf{Acc.}
  & \textbf{F1} & $\Delta$\textbf{F1} \\
\midrule
\rowcolor{mygray}
\yes & \yes & \textbf{Full (Coarse + Fine)}
  & \textbf{86.9} & ---
  & \textbf{86.6} & --- \\
\yes & \no & \;\;\;\;\textit{w/o} Fine
  & 84.9 & \textcolor{myred}{$-2.0$}
  & 84.6 & \textcolor{myred}{$-2.0$} \\
\no & \yes & \;\;\;\;\textit{w/o} Coarse
  & 85.2 & \textcolor{myred}{$-1.7$}
  & 84.6 & \textcolor{myred}{$-2.0$} \\
\no & \no & \;\;\;\;\textit{w/o} Both (image-only)
  & 81.5 & \textcolor{myred}{$-5.4$}
  & 80.7 & \textcolor{myred}{$-5.9$} \\
\bottomrule
\end{tabular}%
}
\caption{
\textbf{Ablation study of rubric construction on Qwen3.5-122B-A10B.}
%We start from the full model (top row) and progressively remove the instruction-specific Fine rubric and the task-family Coarse rubric. The bottom \textit{image-only} row provides no evaluation criterion at all, so the verifier must judge success purely from the instruction and the trajectory screenshots.
$\Delta$Acc.\ and $\Delta$F1 report the absolute drop relative to the full model.
}
\label{tab:ablation_study}
\end{table}

\subsection{Ablation Study}

Table~\ref{tab:ablation_study} isolates the contribution of the Coarse and Fine rubrics in \method.
The full model achieves \textbf{86.9} accuracy and \textbf{86.6} F1 on Qwen3.5-122B-A10B.
Removing either component consistently weakens reward verification.
Without the Fine rubric, F1 drops by \textbf{2.0} points, and without the Coarse rubric, F1 drops by \textbf{2.0} points as well.
The largest decline appears when both rubrics are removed, where the verifier relies only on the instruction and trajectory screenshots and F1 falls by \textbf{5.9} points.
This pattern shows that the two rubrics are complementary, since the Coarse rubric anchors the verifier to task-category boundaries and the Fine rubric adds instruction-specific success requirements.

% \begin{table*}[!t]
%   \centering
%   \scriptsize
%   \setlength{\tabcolsep}{4.2pt}
%   \renewcommand{\arraystretch}{1.12}
%   \begin{tabular}{lrrrrrrrr}
%     \toprule
%     \textbf{Method} &
%     \multicolumn{2}{c}{\textbf{Quality}} &
%     \multicolumn{4}{c}{\textbf{LLM Cost}} &
%     \multicolumn{2}{c}{\textbf{Runtime}} \\
%     \cmidrule(lr){2-3}\cmidrule(lr){4-7}\cmidrule(lr){8-9}
%     & \textbf{Acc.} & \textbf{F1} &
%     \textbf{Calls} & \textbf{Calls / traj.} &
%     \textbf{Tokens} & \textbf{Tokens / traj.} &
%     \textbf{Time} & \textbf{s / traj.} \\
%     \midrule
%     OS-Themis & 79.3 & 77.2 & 21{,}490 & 15.25 & 243.75M & 173.0K & 01:28:47 & 3.78 \\
%     ZeroGUI & 83.0 & 82.6 & 5{,}608 & 3.98 & 100.15M & 71.1K & 00:51:42 & 2.20 \\
%     DigiRL & 74.9 & 74.8 & \textbf{1{,}399} & \textbf{0.99} & \textbf{3.52M} & \textbf{2.5K} & \textbf{00:03:34} & \textbf{0.15} \\
%     \rowcolor{mygray}
%     \textbf{\method} & \textbf{83.4} & \textbf{83.8} & 4{,}220 & 3.00 & 26.98M & 19.1K & 00:36:38 & 1.56 \\
%     \bottomrule
%   \end{tabular}
%   \caption{\textbf{Efficiency comparison on OGRBench using Qwen3-VL-8B-Instruct.}
%   All methods are evaluated on the same 1{,}409 trajectories with 16 parallel workers.
%   Calls count all LLM requests issued by each pipeline, and tokens include prompt and completion tokens.
%   The lowest-cost entries are bolded; \method achieves the best quality while remaining substantially cheaper than the strong multi-call and multi-vote baselines.}
%   \label{tab:efficiency}
% \end{table*}

\begin{table}[t]
  \centering
  \scriptsize
  \setlength{\tabcolsep}{4.4pt}
  \renewcommand{\arraystretch}{1.12}
  \resizebox{0.95\columnwidth}{!}{%
  \begin{tabular}{@{}lcccccccc@{}}
    \toprule
    \multirow{2}{*}{\textbf{Method}} &
    \multicolumn{2}{c}{\textbf{Quality}} &
    \multicolumn{4}{c}{\textbf{LLM Cost}} &
    \multicolumn{2}{c}{\textbf{Runtime}} \\
    \cmidrule(lr){2-3}\cmidrule(lr){4-7}\cmidrule(l){8-9}
    & \textbf{Acc.} & \textbf{F1} &
    \textbf{Calls} & \textbf{Calls/traj.} &
    \textbf{Tokens} & \textbf{Tok./traj.} &
    \textbf{Min.} & \textbf{s/traj.} \\
    \midrule
    OS-Themis & 78.7 & 76.5 & 21{,}490 & 15.25 & 243.75M & 173.0K & 88.8 & 3.78 \\
    ZeroGUI   & 83.3 & 83.0 & 5{,}608  & 3.98  & 100.15M & 71.1K  & 51.7 & 2.20 \\
    DigiRL    & 78.7 & 80.8 & \textbf{1{,}402} & \textbf{1.00} &
    \textbf{23.46M} & \textbf{16.6K} & \textbf{34.2} & \textbf{1.46} \\
    \rowcolor{mygray}
    \textbf{\method} & \textbf{84.0} & \textbf{84.2} &
    4{,}220 & 3.00 & 26.98M & 19.1K & 36.6 & 1.56 \\
    \bottomrule
  \end{tabular}
  }
  \caption{\textbf{Efficiency comparison on OGRBench using Qwen3-VL-8B-Instruct.}
  All methods use the same ten-screenshot trajectory budget. Lowest cost in each column is in bold.}
  \label{tab:efficiency}
\end{table}

\subsection{Efficiency Analysis}

Table~\ref{tab:efficiency} compares OGRBench quality and inference cost using Qwen3-VL-8B-Instruct.
\method~achieves the best accuracy and F1, with \textbf{84.0} Acc.\ and \textbf{84.2} F1, while using substantially fewer calls, tokens, and runtime than OS-Themis and ZeroGUI.
DigiRL is slightly cheaper, but its F1 remains \textbf{3.4} points lower than \method.
Overall, \method~provides the strongest quality--cost trade-off for repeated reward verification.

\subsection{Case Study}

Appendix~\ref{app:case_study} presents a representative failure case showing why task-adaptive criteria matter.
The task asks the agent to add text to the top of a note, but the trajectory inserts the text below existing content.
Baselines focus on whether the requested text appears and incorrectly mark the task as successful.
\method~combines a coarse create/modify rubric with fine placement cues, allowing the verifier to check both the edited item and the instruction-specific ordering constraint.

\section{Conclusion}

We presented \method, a coarse-to-fine rubric framework for task-adaptive GUI outcome reward modeling.
The category-level Coarse rubric anchors verification to task-family success criteria, while the instance-level Fine rubric specifies the task-adaptive criteria needed for the current instruction.
Across offline and online experiments, \method~consistently demonstrates the value of explicitly constructing task-adaptive criteria for success, supporting our core claim for reliable GUI reward modeling across GUI interaction settings.
% Extensive offline and online experiments demonstrate the effectiveness of this design.
% On OGRBench, \method~achieves the best average performance under a matched image budget and improves reward-guided test-time trajectory selection; in online reinforcement learning, it yields the strongest result when used as the reward verifier.
% Ablations further show that both rubric levels contribute to performance, supporting our core claim that reliable GUI reward modeling benefits from explicitly constructing task-adaptive criteria for success.

% \clearpage
\section*{Limitations}
\method~improves GUI reward verification by constructing task-adaptive rubrics, but it still has several scope boundaries. First, although our evaluation covers mobile, desktop, and web environments, it does not exhaust all possible applications, interface designs, or user instruction styles. Second, the coarse rubric bank is constructed offline and kept fixed during evaluation to ensure reproducibility and controlled comparison across methods; this makes the protocol stable, but it may not capture every newly emerging application or domain-specific workflow. % Finally, our experiments use a finite set of VLM backbones, trajectory budgets, and benchmark tasks, so the reported results should be viewed as a representative but incomplete slice of real-world GUI reward modeling.
% \section*{Ethics Statement}
% TODO: write Ethics Statement.

% Bibliography entries for the entire Anthology, followed by custom entries
%\bibliography{anthology,custom}
% Custom bibliography entries only

\bibliography{custom}

@misc{rawles2025androidworlddynamicbenchmarkingenvironment,
      title={AndroidWorld: A Dynamic Benchmarking Environment for Autonomous Agents}, 
      author={Christopher Rawles and Sarah Clinckemaillie and Yifan Chang and Jonathan Waltz and Gabrielle Lau and Marybeth Fair and Alice Li and William Bishop and Wei Li and Folawiyo Campbell-Ajala and Daniel Toyama and Robert Berry and Divya Tyamagundlu and Timothy Lillicrap and Oriana Riva},
      year={2025},
      eprint={2405.14573},
      archivePrefix={arXiv},
      primaryClass={cs.AI},
      url={https://arxiv.org/abs/2405.14573}, 
}

@misc{liu2025infiguir1advancingmultimodalgui,
      title={InfiGUI-R1: Advancing Multimodal GUI Agents from Reactive Actors to Deliberative Reasoners}, 
      author={Yuhang Liu and Pengxiang Li and Congkai Xie and Xavier Hu and Xiaotian Han and Shengyu Zhang and Hongxia Yang and Fei Wu},
      year={2025},
      eprint={2504.14239},
      archivePrefix={arXiv},
      primaryClass={cs.AI},
      url={https://arxiv.org/abs/2504.14239}, 
}

@misc{pan2024autonomousevaluationrefinementdigital,
      title={Autonomous Evaluation and Refinement of Digital Agents}, 
      author={Jiayi Pan and Yichi Zhang and Nicholas Tomlin and Yifei Zhou and Sergey Levine and Alane Suhr},
      year={2024},
      eprint={2404.06474},
      archivePrefix={arXiv},
      primaryClass={cs.AI},
      url={https://arxiv.org/abs/2404.06474}, 
}

@misc{hong2024cogagentvisuallanguagemodel,
      title={CogAgent: A Visual Language Model for GUI Agents}, 
      author={Wenyi Hong and Weihan Wang and Qingsong Lv and Jiazheng Xu and Wenmeng Yu and Junhui Ji and Yan Wang and Zihan Wang and Yuxuan Zhang and Juanzi Li and Bin Xu and Yuxiao Dong and Ming Ding and Jie Tang},
      year={2024},
      eprint={2312.08914},
      archivePrefix={arXiv},
      primaryClass={cs.CV},
      url={https://arxiv.org/abs/2312.08914}, 
}

@misc{zhang2025instructiontuninglargelanguage,
      title={Instruction Tuning for Large Language Models: A Survey}, 
      author={Shengyu Zhang and Linfeng Dong and Xiaoya Li and Sen Zhang and Xiaofei Sun and Shuhe Wang and Jiwei Li and Runyi Hu and Tianwei Zhang and Fei Wu and Guoyin Wang},
      year={2025},
      eprint={2308.10792},
      archivePrefix={arXiv},
      primaryClass={cs.CL},
      url={https://arxiv.org/abs/2308.10792}, 
}

@misc{hu2025osagentssurveymllmbased,
      title={OS Agents: A Survey on MLLM-based Agents for General Computing Devices Use}, 
      author={Xueyu Hu and Tao Xiong and Biao Yi and Zishu Wei and Ruixuan Xiao and Yurun Chen and Jiasheng Ye and Meiling Tao and Xiangxin Zhou and Ziyu Zhao and Yuhuai Li and Shengze Xu and Shenzhi Wang and Xinchen Xu and Shuofei Qiao and Zhaokai Wang and Kun Kuang and Tieyong Zeng and Liang Wang and Jiwei Li and Yuchen Eleanor Jiang and Wangchunshu Zhou and Guoyin Wang and Keting Yin and Zhou Zhao and Hongxia Yang and Fan Wu and Shengyu Zhang and Fei Wu},
      year={2025},
      eprint={2508.04482},
      archivePrefix={arXiv},
      primaryClass={cs.AI},
      url={https://arxiv.org/abs/2508.04482}, 
}

@misc{tang2026clawguiunifiedframeworktraining,
      title={ClawGUI: A Unified Framework for Training, Evaluating, and Deploying GUI Agents}, 
      author={Fei Tang and Zhiqiong Lu and Boxuan Zhang and Weiming Lu and Jun Xiao and Yueting Zhuang and Yongliang Shen},
      year={2026},
      eprint={2604.11784},
      archivePrefix={arXiv},
      primaryClass={cs.LG},
      url={https://arxiv.org/abs/2604.11784}, 
}

@misc{li2026osthemisscalablecriticframework,
      title={OS-Themis: A Scalable Critic Framework for Generalist GUI Rewards}, 
      author={Zehao Li and Zhenyu Wu and Yibo Zhao and Bowen Yang and Jingjing Xie and Zhaoyang Liu and Zhoumianze Liu and Kaiming Jin and Jianze Liang and Zonglin Li and Feng Wu and Bowen Zhou and Zun Wang and Zichen Ding},
      year={2026},
      eprint={2603.19191},
      archivePrefix={arXiv},
      primaryClass={cs.AI},
      url={https://arxiv.org/abs/2603.19191}, 
}

@misc{yang2025zeroguiautomatingonlinegui,
      title={ZeroGUI: Automating Online GUI Learning at Zero Human Cost}, 
      author={Chenyu Yang and Shiqian Su and Shi Liu and Xuan Dong and Yue Yu and Weijie Su and Xuehui Wang and Zhaoyang Liu and Jinguo Zhu and Hao Li and Wenhai Wang and Yu Qiao and Xizhou Zhu and Jifeng Dai},
      year={2025},
      eprint={2505.23762},
      archivePrefix={arXiv},
      primaryClass={cs.AI},
      url={https://arxiv.org/abs/2505.23762}, 
}

@misc{xie2024osworldbenchmarkingmultimodalagents,
      title={OSWorld: Benchmarking Multimodal Agents for Open-Ended Tasks in Real Computer Environments}, 
      author={Tianbao Xie and Danyang Zhang and Jixuan Chen and Xiaochuan Li and Siheng Zhao and Ruisheng Cao and Toh Jing Hua and Zhoujun Cheng and Dongchan Shin and Fangyu Lei and Yitao Liu and Yiheng Xu and Shuyan Zhou and Silvio Savarese and Caiming Xiong and Victor Zhong and Tao Yu},
      year={2024},
      eprint={2404.07972},
      archivePrefix={arXiv},
      primaryClass={cs.AI},
      url={https://arxiv.org/abs/2404.07972}, 
}

@misc{bonatti2024windowsagentarenaevaluating,
      title={Windows Agent Arena: Evaluating Multi-Modal OS Agents at Scale}, 
      author={Rogerio Bonatti and Dan Zhao and Francesco Bonacci and Dillon Dupont and Sara Abdali and Yinheng Li and Yadong Lu and Justin Wagle and Kazuhito Koishida and Arthur Bucker and Lawrence Jang and Zack Hui},
      year={2024},
      eprint={2409.08264},
      archivePrefix={arXiv},
      primaryClass={cs.AI},
      url={https://arxiv.org/abs/2409.08264}, 
}

@article{wang2025mmbenchgui,
  title   = {MMBench-GUI: Hierarchical Multi-Platform Evaluation Framework for GUI Agents},
  author  = {Xuehui Wang and Zhenyu Wu and JingJing Xie and Zichen Ding and Bowen Yang and Zehao Li and Zhaoyang Liu and Qingyun Li and Xuan Dong and Zhe Chen and Weiyun Wang and Xiangyu Zhao and Jixuan Chen and Haodong Duan and Tianbao Xie and Shiqian Su and Chenyu Yang and Yue Yu and Yuan Huang and Yiqian Liu and Xiao Zhang and Xiangyu Yue and Weijie Su and Xizhou Zhu and Wei Shen and Jifeng Dai and Wenhai Wang},
  journal    = {arXiv preprint arXiv:2507.19478},
  year    = {2025}
}

@article{zhou2023webarena,
  title={WebArena: A Realistic Web Environment for Building Autonomous Agents},
  author={Zhou, Shuyan and Xu, Frank F and Zhu, Hao and Zhou, Xuhui and Lo, Robert and Sridhar, Abishek and Cheng, Xianyi and Bisk, Yonatan and Fried, Daniel and Alon, Uri and others},
  journal={arXiv preprint arXiv:2307.13854},
  year={2023}
}

@article{Qwen3-VL,
      title={Qwen3-VL Technical Report}, 
      author={Shuai Bai and Yuxuan Cai and Ruizhe Chen and Keqin Chen and Xionghui Chen and Zesen Cheng and Lianghao Deng and Wei Ding and Chang Gao and Chunjiang Ge and Wenbin Ge and Zhifang Guo and Qidong Huang and Jie Huang and Fei Huang and Binyuan Hui and Shutong Jiang and Zhaohai Li and Mingsheng Li and Mei Li and Kaixin Li and Zicheng Lin and Junyang Lin and Xuejing Liu and Jiawei Liu and Chenglong Liu and Yang Liu and Dayiheng Liu and Shixuan Liu and Dunjie Lu and Ruilin Luo and Chenxu Lv and Rui Men and Lingchen Meng and Xuancheng Ren and Xingzhang Ren and Sibo Song and Yuchong Sun and Jun Tang and Jianhong Tu and Jianqiang Wan and Peng Wang and Pengfei Wang and Qiuyue Wang and Yuxuan Wang and Tianbao Xie and Yiheng Xu and Haiyang Xu and Jin Xu and Zhibo Yang and Mingkun Yang and Jianxin Yang and An Yang and Bowen Yu and Fei Zhang and Hang Zhang and Xi Zhang and Bo Zheng and Humen Zhong and Jingren Zhou and Fan Zhou and Jing Zhou and Yuanzhi Zhu and Ke Zhu},
	  journal={arXiv preprint arXiv:2511.21631},
      year={2025}
}

@misc{qwen35blog,
    title = {Qwen3.5: Accelerating Productivity with Native Multimodal Agents},
    url = {https://qwen.ai/blog?id=qwen3.5},
    author = {Qwen Team},
    month = {February},
    year = {2026}
}

@misc{qwen3.6-27b,
    title = {{Qwen3.6-27B}: Flagship-Level Coding in a {27B} Dense Model},
    author = {{Qwen Team}},
    year = {2026},
    month = {April},
    url = {https://qwen.ai/blog?id=qwen3.6-27b}
}

@misc{geminiteam2025geminifamilyhighlycapable,
      title={Gemini: A Family of Highly Capable Multimodal Models}, 
      author={Gemini Team and Rohan Anil and Sebastian Borgeaud and Jean-Baptiste Alayrac and Jiahui Yu and Radu Soricut and Johan Schalkwyk and Andrew M. Dai and Anja Hauth and Katie Millican and David Silver and Melvin Johnson and Ioannis Antonoglou and Julian Schrittwieser and Amelia Glaese and Jilin Chen and Emily Pitler and Timothy Lillicrap and Angeliki Lazaridou and Orhan Firat and James Molloy and Michael Isard and Paul R. Barham and Tom Hennigan and Benjamin Lee and Fabio Viola and Malcolm Reynolds and Yuanzhong Xu and Ryan Doherty and Eli Collins and Clemens Meyer and Eliza Rutherford and Erica Moreira and Kareem Ayoub and Megha Goel and Jack Krawczyk and Cosmo Du and Ed Chi and Heng-Tze Cheng and Eric Ni and Purvi Shah and Patrick Kane and Betty Chan and Manaal Faruqui and Aliaksei Severyn and Hanzhao Lin and YaGuang Li and Yong Cheng and Abe Ittycheriah and Mahdis Mahdieh and Mia Chen and Pei Sun and Dustin Tran and Sumit Bagri and Balaji Lakshminarayanan and Jeremiah Liu and Andras Orban and Fabian Güra and Hao Zhou and Xinying Song and Aurelien Boffy and Harish Ganapathy and Steven Zheng and HyunJeong Choe and Ágoston Weisz and Tao Zhu and Yifeng Lu and Siddharth Gopal and Jarrod Kahn and Maciej Kula and Jeff Pitman and Rushin Shah and Emanuel Taropa and Majd Al Merey and Martin Baeuml and Zhifeng Chen and Laurent El Shafey and Yujing Zhang and Olcan Sercinoglu and George Tucker and Enrique Piqueras and Maxim Krikun and Iain Barr and Nikolay Savinov and Ivo Danihelka and Becca Roelofs and Anaïs White and Anders Andreassen and Tamara von Glehn and Lakshman Yagati and Mehran Kazemi and Lucas Gonzalez and Misha Khalman and Jakub Sygnowski and Alexandre Frechette and Charlotte Smith and Laura Culp and Lev Proleev and Yi Luan and Xi Chen and James Lottes and Nathan Schucher and Federico Lebron and Alban Rrustemi and Natalie Clay and Phil Crone and Tomas Kocisky and Jeffrey Zhao and Bartek Perz and Dian Yu and Heidi Howard and Adam Bloniarz and Jack W. Rae and Han Lu and Laurent Sifre and Marcello Maggioni and Fred Alcober and Dan Garrette and Megan Barnes and Shantanu Thakoor and Jacob Austin and Gabriel Barth-Maron and William Wong and Rishabh Joshi and Rahma Chaabouni and Deeni Fatiha and Arun Ahuja and Gaurav Singh Tomar and Evan Senter and Martin Chadwick and Ilya Kornakov and Nithya Attaluri and Iñaki Iturrate and Ruibo Liu and Yunxuan Li and Sarah Cogan and Jeremy Chen and Chao Jia and Chenjie Gu and Qiao Zhang and Jordan Grimstad and Ale Jakse Hartman and Xavier Garcia and Thanumalayan Sankaranarayana Pillai and Jacob Devlin and Michael Laskin and Diego de Las Casas and Dasha Valter and Connie Tao and Lorenzo Blanco and Adrià Puigdomènech Badia and David Reitter and Mianna Chen and Jenny Brennan and Clara Rivera and Sergey Brin and Shariq Iqbal and Gabriela Surita and Jane Labanowski and Abhi Rao and Stephanie Winkler and Emilio Parisotto and Yiming Gu and Kate Olszewska and Ravi Addanki and Antoine Miech and Annie Louis and Denis Teplyashin and Geoff Brown and Elliot Catt and Jan Balaguer and Jackie Xiang and Pidong Wang and Zoe Ashwood and Anton Briukhov and Albert Webson and Sanjay Ganapathy and Smit Sanghavi and Ajay Kannan and Ming-Wei Chang and Axel Stjerngren and Josip Djolonga and Yuting Sun and Ankur Bapna and Matthew Aitchison and Pedram Pejman and Henryk Michalewski and Tianhe Yu and Cindy Wang and Juliette Love and Junwhan Ahn and Dawn Bloxwich and Kehang Han and Peter Humphreys and Thibault Sellam and James Bradbury and Varun Godbole and Sina Samangooei and Bogdan Damoc and Alex Kaskasoli and Sébastien M. R. Arnold and Vijay Vasudevan and Shubham Agrawal and Jason Riesa and Dmitry Lepikhin and Richard Tanburn and Srivatsan Srinivasan and Hyeontaek Lim and Sarah Hodkinson and Pranav Shyam and Johan Ferret and Steven Hand and Ankush Garg and Tom Le Paine and Jian Li and Yujia Li and Minh Giang and Alexander Neitz and Zaheer Abbas and Sarah York and Machel Reid and Elizabeth Cole and Aakanksha Chowdhery and Dipanjan Das and Dominika Rogozińska and Vitaliy Nikolaev and Pablo Sprechmann and Zachary Nado and Lukas Zilka and Flavien Prost and Luheng He and Marianne Monteiro and Gaurav Mishra and Chris Welty and Josh Newlan and Dawei Jia and Miltiadis Allamanis and Clara Huiyi Hu and Raoul de Liedekerke and Justin Gilmer and Carl Saroufim and Shruti Rijhwani and Shaobo Hou and Disha Shrivastava and Anirudh Baddepudi and Alex Goldin and Adnan Ozturel and Albin Cassirer and Yunhan Xu and Daniel Sohn and Devendra Sachan and Reinald Kim Amplayo and Craig Swanson and Dessie Petrova and Shashi Narayan and Arthur Guez and Siddhartha Brahma and Jessica Landon and Miteyan Patel and Ruizhe Zhao and Kevin Villela and Luyu Wang and Wenhao Jia and Matthew Rahtz and Mai Giménez and Legg Yeung and James Keeling and Petko Georgiev and Diana Mincu and Boxi Wu and Salem Haykal and Rachel Saputro and Kiran Vodrahalli and James Qin and Zeynep Cankara and Abhanshu Sharma and Nick Fernando and Will Hawkins and Behnam Neyshabur and Solomon Kim and Adrian Hutter and Priyanka Agrawal and Alex Castro-Ros and George van den Driessche and Tao Wang and Fan Yang and Shuo-yiin Chang and Paul Komarek and Ross McIlroy and Mario Lučić and Guodong Zhang and Wael Farhan and Michael Sharman and Paul Natsev and Paul Michel and Yamini Bansal and Siyuan Qiao and Kris Cao and Siamak Shakeri and Christina Butterfield and Justin Chung and Paul Kishan Rubenstein and Shivani Agrawal and Arthur Mensch and Kedar Soparkar and Karel Lenc and Timothy Chung and Aedan Pope and Loren Maggiore and Jackie Kay and Priya Jhakra and Shibo Wang and Joshua Maynez and Mary Phuong and Taylor Tobin and Andrea Tacchetti and Maja Trebacz and Kevin Robinson and Yash Katariya and Sebastian Riedel and Paige Bailey and Kefan Xiao and Nimesh Ghelani and Lora Aroyo and Ambrose Slone and Neil Houlsby and Xuehan Xiong and Zhen Yang and Elena Gribovskaya and Jonas Adler and Mateo Wirth and Lisa Lee and Music Li and Thais Kagohara and Jay Pavagadhi and Sophie Bridgers and Anna Bortsova and Sanjay Ghemawat and Zafarali Ahmed and Tianqi Liu and Richard Powell and Vijay Bolina and Mariko Iinuma and Polina Zablotskaia and James Besley and Da-Woon Chung and Timothy Dozat and Ramona Comanescu and Xiance Si and Jeremy Greer and Guolong Su and Martin Polacek and Raphaël Lopez Kaufman and Simon Tokumine and Hexiang Hu and Elena Buchatskaya and Yingjie Miao and Mohamed Elhawaty and Aditya Siddhant and Nenad Tomasev and Jinwei Xing and Christina Greer and Helen Miller and Shereen Ashraf and Aurko Roy and Zizhao Zhang and Ada Ma and Angelos Filos and Milos Besta and Rory Blevins and Ted Klimenko and Chih-Kuan Yeh and Soravit Changpinyo and Jiaqi Mu and Oscar Chang and Mantas Pajarskas and Carrie Muir and Vered Cohen and Charline Le Lan and Krishna Haridasan and Amit Marathe and Steven Hansen and Sholto Douglas and Rajkumar Samuel and Mingqiu Wang and Sophia Austin and Chang Lan and Jiepu Jiang and Justin Chiu and Jaime Alonso Lorenzo and Lars Lowe Sjösund and Sébastien Cevey and Zach Gleicher and Thi Avrahami and Anudhyan Boral and Hansa Srinivasan and Vittorio Selo and Rhys May and Konstantinos Aisopos and Léonard Hussenot and Livio Baldini Soares and Kate Baumli and Michael B. Chang and Adrià Recasens and Ben Caine and Alexander Pritzel and Filip Pavetic and Fabio Pardo and Anita Gergely and Justin Frye and Vinay Ramasesh and Dan Horgan and Kartikeya Badola and Nora Kassner and Subhrajit Roy and Ethan Dyer and Víctor Campos Campos and Alex Tomala and Yunhao Tang and Dalia El Badawy and Elspeth White and Basil Mustafa and Oran Lang and Abhishek Jindal and Sharad Vikram and Zhitao Gong and Sergi Caelles and Ross Hemsley and Gregory Thornton and Fangxiaoyu Feng and Wojciech Stokowiec and Ce Zheng and Phoebe Thacker and Çağlar Ünlü and Zhishuai Zhang and Mohammad Saleh and James Svensson and Max Bileschi and Piyush Patil and Ankesh Anand and Roman Ring and Katerina Tsihlas and Arpi Vezer and Marco Selvi and Toby Shevlane and Mikel Rodriguez and Tom Kwiatkowski and Samira Daruki and Keran Rong and Allan Dafoe and Nicholas FitzGerald and Keren Gu-Lemberg and Mina Khan and Lisa Anne Hendricks and Marie Pellat and Vladimir Feinberg and James Cobon-Kerr and Tara Sainath and Maribeth Rauh and Sayed Hadi Hashemi and Richard Ives and Yana Hasson and Eric Noland and Yuan Cao and Nathan Byrd and Le Hou and Qingze Wang and Thibault Sottiaux and Michela Paganini and Jean-Baptiste Lespiau and Alexandre Moufarek and Samer Hassan and Kaushik Shivakumar and Joost van Amersfoort and Amol Mandhane and Pratik Joshi and Anirudh Goyal and Matthew Tung and Andrew Brock and Hannah Sheahan and Vedant Misra and Cheng Li and Nemanja Rakićević and Mostafa Dehghani and Fangyu Liu and Sid Mittal and Junhyuk Oh and Seb Noury and Eren Sezener and Fantine Huot and Matthew Lamm and Nicola De Cao and Charlie Chen and Sidharth Mudgal and Romina Stella and Kevin Brooks and Gautam Vasudevan and Chenxi Liu and Mainak Chain and Nivedita Melinkeri and Aaron Cohen and Venus Wang and Kristie Seymore and Sergey Zubkov and Rahul Goel and Summer Yue and Sai Krishnakumaran and Brian Albert and Nate Hurley and Motoki Sano and Anhad Mohananey and Jonah Joughin and Egor Filonov and Tomasz Kępa and Yomna Eldawy and Jiawern Lim and Rahul Rishi and Shirin Badiezadegan and Taylor Bos and Jerry Chang and Sanil Jain and Sri Gayatri Sundara Padmanabhan and Subha Puttagunta and Kalpesh Krishna and Leslie Baker and Norbert Kalb and Vamsi Bedapudi and Adam Kurzrok and Shuntong Lei and Anthony Yu and Oren Litvin and Xiang Zhou and Zhichun Wu and Sam Sobell and Andrea Siciliano and Alan Papir and Robby Neale and Jonas Bragagnolo and Tej Toor and Tina Chen and Valentin Anklin and Feiran Wang and Richie Feng and Milad Gholami and Kevin Ling and Lijuan Liu and Jules Walter and Hamid Moghaddam and Arun Kishore and Jakub Adamek and Tyler Mercado and Jonathan Mallinson and Siddhinita Wandekar and Stephen Cagle and Eran Ofek and Guillermo Garrido and Clemens Lombriser and Maksim Mukha and Botu Sun and Hafeezul Rahman Mohammad and Josip Matak and Yadi Qian and Vikas Peswani and Pawel Janus and Quan Yuan and Leif Schelin and Oana David and Ankur Garg and Yifan He and Oleksii Duzhyi and Anton Älgmyr and Timothée Lottaz and Qi Li and Vikas Yadav and Luyao Xu and Alex Chinien and Rakesh Shivanna and Aleksandr Chuklin and Josie Li and Carrie Spadine and Travis Wolfe and Kareem Mohamed and Subhabrata Das and Zihang Dai and Kyle He and Daniel von Dincklage and Shyam Upadhyay and Akanksha Maurya and Luyan Chi and Sebastian Krause and Khalid Salama and Pam G Rabinovitch and Pavan Kumar Reddy M and Aarush Selvan and Mikhail Dektiarev and Golnaz Ghiasi and Erdem Guven and Himanshu Gupta and Boyi Liu and Deepak Sharma and Idan Heimlich Shtacher and Shachi Paul and Oscar Akerlund and François-Xavier Aubet and Terry Huang and Chen Zhu and Eric Zhu and Elico Teixeira and Matthew Fritze and Francesco Bertolini and Liana-Eleonora Marinescu and Martin Bölle and Dominik Paulus and Khyatti Gupta and Tejasi Latkar and Max Chang and Jason Sanders and Roopa Wilson and Xuewei Wu and Yi-Xuan Tan and Lam Nguyen Thiet and Tulsee Doshi and Sid Lall and Swaroop Mishra and Wanming Chen and Thang Luong and Seth Benjamin and Jasmine Lee and Ewa Andrejczuk and Dominik Rabiej and Vipul Ranjan and Krzysztof Styrc and Pengcheng Yin and Jon Simon and Malcolm Rose Harriott and Mudit Bansal and Alexei Robsky and Geoff Bacon and David Greene and Daniil Mirylenka and Chen Zhou and Obaid Sarvana and Abhimanyu Goyal and Samuel Andermatt and Patrick Siegler and Ben Horn and Assaf Israel and Francesco Pongetti and Chih-Wei "Louis" Chen and Marco Selvatici and Pedro Silva and Kathie Wang and Jackson Tolins and Kelvin Guu and Roey Yogev and Xiaochen Cai and Alessandro Agostini and Maulik Shah and Hung Nguyen and Noah Ó Donnaile and Sébastien Pereira and Linda Friso and Adam Stambler and Adam Kurzrok and Chenkai Kuang and Yan Romanikhin and Mark Geller and ZJ Yan and Kane Jang and Cheng-Chun Lee and Wojciech Fica and Eric Malmi and Qijun Tan and Dan Banica and Daniel Balle and Ryan Pham and Yanping Huang and Diana Avram and Hongzhi Shi and Jasjot Singh and Chris Hidey and Niharika Ahuja and Pranab Saxena and Dan Dooley and Srividya Pranavi Potharaju and Eileen O'Neill and Anand Gokulchandran and Ryan Foley and Kai Zhao and Mike Dusenberry and Yuan Liu and Pulkit Mehta and Ragha Kotikalapudi and Chalence Safranek-Shrader and Andrew Goodman and Joshua Kessinger and Eran Globen and Prateek Kolhar and Chris Gorgolewski and Ali Ibrahim and Yang Song and Ali Eichenbaum and Thomas Brovelli and Sahitya Potluri and Preethi Lahoti and Cip Baetu and Ali Ghorbani and Charles Chen and Andy Crawford and Shalini Pal and Mukund Sridhar and Petru Gurita and Asier Mujika and Igor Petrovski and Pierre-Louis Cedoz and Chenmei Li and Shiyuan Chen and Niccolò Dal Santo and Siddharth Goyal and Jitesh Punjabi and Karthik Kappaganthu and Chester Kwak and Pallavi LV and Sarmishta Velury and Himadri Choudhury and Jamie Hall and Premal Shah and Ricardo Figueira and Matt Thomas and Minjie Lu and Ting Zhou and Chintu Kumar and Thomas Jurdi and Sharat Chikkerur and Yenai Ma and Adams Yu and Soo Kwak and Victor Ähdel and Sujeevan Rajayogam and Travis Choma and Fei Liu and Aditya Barua and Colin Ji and Ji Ho Park and Vincent Hellendoorn and Alex Bailey and Taylan Bilal and Huanjie Zhou and Mehrdad Khatir and Charles Sutton and Wojciech Rzadkowski and Fiona Macintosh and Roopali Vij and Konstantin Shagin and Paul Medina and Chen Liang and Jinjing Zhou and Pararth Shah and Yingying Bi and Attila Dankovics and Shipra Banga and Sabine Lehmann and Marissa Bredesen and Zifan Lin and John Eric Hoffmann and Jonathan Lai and Raynald Chung and Kai Yang and Nihal Balani and Arthur Bražinskas and Andrei Sozanschi and Matthew Hayes and Héctor Fernández Alcalde and Peter Makarov and Will Chen and Antonio Stella and Liselotte Snijders and Michael Mandl and Ante Kärrman and Paweł Nowak and Xinyi Wu and Alex Dyck and Krishnan Vaidyanathan and Raghavender R and Jessica Mallet and Mitch Rudominer and Eric Johnston and Sushil Mittal and Akhil Udathu and Janara Christensen and Vishal Verma and Zach Irving and Andreas Santucci and Gamaleldin Elsayed and Elnaz Davoodi and Marin Georgiev and Ian Tenney and Nan Hua and Geoffrey Cideron and Edouard Leurent and Mahmoud Alnahlawi and Ionut Georgescu and Nan Wei and Ivy Zheng and Dylan Scandinaro and Heinrich Jiang and Jasper Snoek and Mukund Sundararajan and Xuezhi Wang and Zack Ontiveros and Itay Karo and Jeremy Cole and Vinu Rajashekhar and Lara Tumeh and Eyal Ben-David and Rishub Jain and Jonathan Uesato and Romina Datta and Oskar Bunyan and Shimu Wu and John Zhang and Piotr Stanczyk and Ye Zhang and David Steiner and Subhajit Naskar and Michael Azzam and Matthew Johnson and Adam Paszke and Chung-Cheng Chiu and Jaume Sanchez Elias and Afroz Mohiuddin and Faizan Muhammad and Jin Miao and Andrew Lee and Nino Vieillard and Jane Park and Jiageng Zhang and Jeff Stanway and Drew Garmon and Abhijit Karmarkar and Zhe Dong and Jong Lee and Aviral Kumar and Luowei Zhou and Jonathan Evens and William Isaac and Geoffrey Irving and Edward Loper and Michael Fink and Isha Arkatkar and Nanxin Chen and Izhak Shafran and Ivan Petrychenko and Zhe Chen and Johnson Jia and Anselm Levskaya and Zhenkai Zhu and Peter Grabowski and Yu Mao and Alberto Magni and Kaisheng Yao and Javier Snaider and Norman Casagrande and Evan Palmer and Paul Suganthan and Alfonso Castaño and Irene Giannoumis and Wooyeol Kim and Mikołaj Rybiński and Ashwin Sreevatsa and Jennifer Prendki and David Soergel and Adrian Goedeckemeyer and Willi Gierke and Mohsen Jafari and Meenu Gaba and Jeremy Wiesner and Diana Gage Wright and Yawen Wei and Harsha Vashisht and Yana Kulizhskaya and Jay Hoover and Maigo Le and Lu Li and Chimezie Iwuanyanwu and Lu Liu and Kevin Ramirez and Andrey Khorlin and Albert Cui and Tian LIN and Marcus Wu and Ricardo Aguilar and Keith Pallo and Abhishek Chakladar and Ginger Perng and Elena Allica Abellan and Mingyang Zhang and Ishita Dasgupta and Nate Kushman and Ivo Penchev and Alena Repina and Xihui Wu and Tom van der Weide and Priya Ponnapalli and Caroline Kaplan and Jiri Simsa and Shuangfeng Li and Olivier Dousse and Fan Yang and Jeff Piper and Nathan Ie and Rama Pasumarthi and Nathan Lintz and Anitha Vijayakumar and Daniel Andor and Pedro Valenzuela and Minnie Lui and Cosmin Paduraru and Daiyi Peng and Katherine Lee and Shuyuan Zhang and Somer Greene and Duc Dung Nguyen and Paula Kurylowicz and Cassidy Hardin and Lucas Dixon and Lili Janzer and Kiam Choo and Ziqiang Feng and Biao Zhang and Achintya Singhal and Dayou Du and Dan McKinnon and Natasha Antropova and Tolga Bolukbasi and Orgad Keller and David Reid and Daniel Finchelstein and Maria Abi Raad and Remi Crocker and Peter Hawkins and Robert Dadashi and Colin Gaffney and Ken Franko and Anna Bulanova and Rémi Leblond and Shirley Chung and Harry Askham and Luis C. Cobo and Kelvin Xu and Felix Fischer and Jun Xu and Christina Sorokin and Chris Alberti and Chu-Cheng Lin and Colin Evans and Alek Dimitriev and Hannah Forbes and Dylan Banarse and Zora Tung and Mark Omernick and Colton Bishop and Rachel Sterneck and Rohan Jain and Jiawei Xia and Ehsan Amid and Francesco Piccinno and Xingyu Wang and Praseem Banzal and Daniel J. Mankowitz and Alex Polozov and Victoria Krakovna and Sasha Brown and MohammadHossein Bateni and Dennis Duan and Vlad Firoiu and Meghana Thotakuri and Tom Natan and Matthieu Geist and Ser tan Girgin and Hui Li and Jiayu Ye and Ofir Roval and Reiko Tojo and Michael Kwong and James Lee-Thorp and Christopher Yew and Danila Sinopalnikov and Sabela Ramos and John Mellor and Abhishek Sharma and Kathy Wu and David Miller and Nicolas Sonnerat and Denis Vnukov and Rory Greig and Jennifer Beattie and Emily Caveness and Libin Bai and Julian Eisenschlos and Alex Korchemniy and Tomy Tsai and Mimi Jasarevic and Weize Kong and Phuong Dao and Zeyu Zheng and Frederick Liu and Fan Yang and Rui Zhu and Tian Huey Teh and Jason Sanmiya and Evgeny Gladchenko and Nejc Trdin and Daniel Toyama and Evan Rosen and Sasan Tavakkol and Linting Xue and Chen Elkind and Oliver Woodman and John Carpenter and George Papamakarios and Rupert Kemp and Sushant Kafle and Tanya Grunina and Rishika Sinha and Alice Talbert and Diane Wu and Denese Owusu-Afriyie and Cosmo Du and Chloe Thornton and Jordi Pont-Tuset and Pradyumna Narayana and Jing Li and Saaber Fatehi and John Wieting and Omar Ajmeri and Benigno Uria and Yeongil Ko and Laura Knight and Amélie Héliou and Ning Niu and Shane Gu and Chenxi Pang and Yeqing Li and Nir Levine and Ariel Stolovich and Rebeca Santamaria-Fernandez and Sonam Goenka and Wenny Yustalim and Robin Strudel and Ali Elqursh and Charlie Deck and Hyo Lee and Zonglin Li and Kyle Levin and Raphael Hoffmann and Dan Holtmann-Rice and Olivier Bachem and Sho Arora and Christy Koh and Soheil Hassas Yeganeh and Siim Põder and Mukarram Tariq and Yanhua Sun and Lucian Ionita and Mojtaba Seyedhosseini and Pouya Tafti and Zhiyu Liu and Anmol Gulati and Jasmine Liu and Xinyu Ye and Bart Chrzaszcz and Lily Wang and Nikhil Sethi and Tianrun Li and Ben Brown and Shreya Singh and Wei Fan and Aaron Parisi and Joe Stanton and Vinod Koverkathu and Christopher A. Choquette-Choo and Yunjie Li and TJ Lu and Abe Ittycheriah and Prakash Shroff and Mani Varadarajan and Sanaz Bahargam and Rob Willoughby and David Gaddy and Guillaume Desjardins and Marco Cornero and Brona Robenek and Bhavishya Mittal and Ben Albrecht and Ashish Shenoy and Fedor Moiseev and Henrik Jacobsson and Alireza Ghaffarkhah and Morgane Rivière and Alanna Walton and Clément Crepy and Alicia Parrish and Zongwei Zhou and Clement Farabet and Carey Radebaugh and Praveen Srinivasan and Claudia van der Salm and Andreas Fidjeland and Salvatore Scellato and Eri Latorre-Chimoto and Hanna Klimczak-Plucińska and David Bridson and Dario de Cesare and Tom Hudson and Piermaria Mendolicchio and Lexi Walker and Alex Morris and Matthew Mauger and Alexey Guseynov and Alison Reid and Seth Odoom and Lucia Loher and Victor Cotruta and Madhavi Yenugula and Dominik Grewe and Anastasia Petrushkina and Tom Duerig and Antonio Sanchez and Steve Yadlowsky and Amy Shen and Amir Globerson and Lynette Webb and Sahil Dua and Dong Li and Surya Bhupatiraju and Dan Hurt and Haroon Qureshi and Ananth Agarwal and Tomer Shani and Matan Eyal and Anuj Khare and Shreyas Rammohan Belle and Lei Wang and Chetan Tekur and Mihir Sanjay Kale and Jinliang Wei and Ruoxin Sang and Brennan Saeta and Tyler Liechty and Yi Sun and Yao Zhao and Stephan Lee and Pandu Nayak and Doug Fritz and Manish Reddy Vuyyuru and John Aslanides and Nidhi Vyas and Martin Wicke and Xiao Ma and Evgenii Eltyshev and Nina Martin and Hardie Cate and James Manyika and Keyvan Amiri and Yelin Kim and Xi Xiong and Kai Kang and Florian Luisier and Nilesh Tripuraneni and David Madras and Mandy Guo and Austin Waters and Oliver Wang and Joshua Ainslie and Jason Baldridge and Han Zhang and Garima Pruthi and Jakob Bauer and Feng Yang and Riham Mansour and Jason Gelman and Yang Xu and George Polovets and Ji Liu and Honglong Cai and Warren Chen and XiangHai Sheng and Emily Xue and Sherjil Ozair and Christof Angermueller and Xiaowei Li and Anoop Sinha and Weiren Wang and Julia Wiesinger and Emmanouil Koukoumidis and Yuan Tian and Anand Iyer and Madhu Gurumurthy and Mark Goldenson and Parashar Shah and MK Blake and Hongkun Yu and Anthony Urbanowicz and Jennimaria Palomaki and Chrisantha Fernando and Ken Durden and Harsh Mehta and Nikola Momchev and Elahe Rahimtoroghi and Maria Georgaki and Amit Raul and Sebastian Ruder and Morgan Redshaw and Jinhyuk Lee and Denny Zhou and Komal Jalan and Dinghua Li and Blake Hechtman and Parker Schuh and Milad Nasr and Kieran Milan and Vladimir Mikulik and Juliana Franco and Tim Green and Nam Nguyen and Joe Kelley and Aroma Mahendru and Andrea Hu and Joshua Howland and Ben Vargas and Jeffrey Hui and Kshitij Bansal and Vikram Rao and Rakesh Ghiya and Emma Wang and Ke Ye and Jean Michel Sarr and Melanie Moranski Preston and Madeleine Elish and Steve Li and Aakash Kaku and Jigar Gupta and Ice Pasupat and Da-Cheng Juan and Milan Someswar and Tejvi M. and Xinyun Chen and Aida Amini and Alex Fabrikant and Eric Chu and Xuanyi Dong and Amruta Muthal and Senaka Buthpitiya and Sarthak Jauhari and Nan Hua and Urvashi Khandelwal and Ayal Hitron and Jie Ren and Larissa Rinaldi and Shahar Drath and Avigail Dabush and Nan-Jiang Jiang and Harshal Godhia and Uli Sachs and Anthony Chen and Yicheng Fan and Hagai Taitelbaum and Hila Noga and Zhuyun Dai and James Wang and Chen Liang and Jenny Hamer and Chun-Sung Ferng and Chenel Elkind and Aviel Atias and Paulina Lee and Vít Listík and Mathias Carlen and Jan van de Kerkhof and Marcin Pikus and Krunoslav Zaher and Paul Müller and Sasha Zykova and Richard Stefanec and Vitaly Gatsko and Christoph Hirnschall and Ashwin Sethi and Xingyu Federico Xu and Chetan Ahuja and Beth Tsai and Anca Stefanoiu and Bo Feng and Keshav Dhandhania and Manish Katyal and Akshay Gupta and Atharva Parulekar and Divya Pitta and Jing Zhao and Vivaan Bhatia and Yashodha Bhavnani and Omar Alhadlaq and Xiaolin Li and Peter Danenberg and Dennis Tu and Alex Pine and Vera Filippova and Abhipso Ghosh and Ben Limonchik and Bhargava Urala and Chaitanya Krishna Lanka and Derik Clive and Yi Sun and Edward Li and Hao Wu and Kevin Hongtongsak and Ianna Li and Kalind Thakkar and Kuanysh Omarov and Kushal Majmundar and Michael Alverson and Michael Kucharski and Mohak Patel and Mudit Jain and Maksim Zabelin and Paolo Pelagatti and Rohan Kohli and Saurabh Kumar and Joseph Kim and Swetha Sankar and Vineet Shah and Lakshmi Ramachandruni and Xiangkai Zeng and Ben Bariach and Laura Weidinger and Tu Vu and Alek Andreev and Antoine He and Kevin Hui and Sheleem Kashem and Amar Subramanya and Sissie Hsiao and Demis Hassabis and Koray Kavukcuoglu and Adam Sadovsky and Quoc Le and Trevor Strohman and Yonghui Wu and Slav Petrov and Jeffrey Dean and Oriol Vinyals},
      year={2025},
      eprint={2312.11805},
      archivePrefix={arXiv},
      primaryClass={cs.CL},
      url={https://arxiv.org/abs/2312.11805}, 
}

@misc{kong2025mobileworldbenchmarkingautonomousmobile,
      title={MobileWorld: Benchmarking Autonomous Mobile Agents in Agent-User Interactive and MCP-Augmented Environments}, 
      author={Quyu Kong and Xu Zhang and Zhenyu Yang and Nolan Gao and Chen Liu and Panrong Tong and Chenglin Cai and Hanzhang Zhou and Jianan Zhang and Liangyu Chen and Zhidan Liu and Steven Hoi and Yue Wang},
      year={2025},
      eprint={2512.19432},
      archivePrefix={arXiv},
      primaryClass={cs.CL},
      url={https://arxiv.org/abs/2512.19432}, 
}

@misc{zhou2025maiuitechnicalreportrealworld,
      title={MAI-UI Technical Report: Real-World Centric Foundation GUI Agents}, 
      author={Hanzhang Zhou and Xu Zhang and Panrong Tong and Jianan Zhang and Liangyu Chen and Quyu Kong and Chenglin Cai and Chen Liu and Yue Wang and Jingren Zhou and Steven Hoi},
      year={2025},
      eprint={2512.22047},
      archivePrefix={arXiv},
      primaryClass={cs.CV},
      url={https://arxiv.org/abs/2512.22047}, 
}

@misc{shao2024deepseekmathpushinglimitsmathematical,
      title={DeepSeekMath: Pushing the Limits of Mathematical Reasoning in Open Language Models}, 
      author={Zhihong Shao and Peiyi Wang and Qihao Zhu and Runxin Xu and Junxiao Song and Xiao Bi and Haowei Zhang and Mingchuan Zhang and Y. K. Li and Y. Wu and Daya Guo},
      year={2024},
      eprint={2402.03300},
      archivePrefix={arXiv},
      primaryClass={cs.CL},
      url={https://arxiv.org/abs/2402.03300}, 
}

@misc{bai2024digirltraininginthewilddevicecontrol,
      title={DigiRL: Training In-The-Wild Device-Control Agents with Autonomous Reinforcement Learning}, 
      author={Hao Bai and Yifei Zhou and Mert Cemri and Jiayi Pan and Alane Suhr and Sergey Levine and Aviral Kumar},
      year={2024},
      eprint={2406.11896},
      archivePrefix={arXiv},
      primaryClass={cs.LG},
      url={https://arxiv.org/abs/2406.11896}, 
}

@misc{liu2025infiguiagentmultimodalgeneralistgui,
      title={InfiGUIAgent: A Multimodal Generalist GUI Agent with Native Reasoning and Reflection}, 
      author={Yuhang Liu and Pengxiang Li and Zishu Wei and Congkai Xie and Xueyu Hu and Xinchen Xu and Shengyu Zhang and Xiaotian Han and Hongxia Yang and Fei Wu},
      year={2025},
      eprint={2501.04575},
      archivePrefix={arXiv},
      primaryClass={cs.AI},
      url={https://arxiv.org/abs/2501.04575}, 
}

@article{wang2024gui,
  title={Gui agents with foundation models: A comprehensive survey},
  author={Wang, Shuai and Liu, Weiwen and Chen, Jingxuan and Zhou, Yuqi and Gan, Weinan and Zeng, Xingshan and Che, Yuhan and Yu, Shuai and Hao, Xinlong and Shao, Kun and others},
  journal={arXiv preprint arXiv:2411.04890},
  year={2024}
}

@misc{xu2025mobilerlonlineagenticreinforcement,
      title={MobileRL: Online Agentic Reinforcement Learning for Mobile GUI Agents}, 
      author={Yifan Xu and Xiao Liu and Xinghan Liu and Jiaqi Fu and Hanchen Zhang and Bohao Jing and Shudan Zhang and Yuting Wang and Wenyi Zhao and Yuxiao Dong},
      year={2025},
      eprint={2509.18119},
      archivePrefix={arXiv},
      primaryClass={cs.LG},
      url={https://arxiv.org/abs/2509.18119}, 
}

@misc{xu2026mobileagentv35multiplatformfundamentalgui,
      title={Mobile-Agent-v3.5: Multi-platform Fundamental GUI Agents}, 
      author={Haiyang Xu and Xi Zhang and Haowei Liu and Junyang Wang and Zhaozai Zhu and Shengjie Zhou and Xuhao Hu and Feiyu Gao and Junjie Cao and Zihua Wang and Zhiyuan Chen and Jitong Liao and Qi Zheng and Jiahui Zeng and Ze Xu and Shuai Bai and Junyang Lin and Jingren Zhou and Ming Yan},
      year={2026},
      eprint={2602.16855},
      archivePrefix={arXiv},
      primaryClass={cs.AI},
      url={https://arxiv.org/abs/2602.16855}, 
}

@misc{xia2025agentrmenhancingagentgeneralization,
      title={AgentRM: Enhancing Agent Generalization with Reward Modeling}, 
      author={Yu Xia and Jingru Fan and Weize Chen and Siyu Yan and Xin Cong and Zhong Zhang and Yaxi Lu and Yankai Lin and Zhiyuan Liu and Maosong Sun},
      year={2025},
      eprint={2502.18407},
      archivePrefix={arXiv},
      primaryClass={cs.CL},
      url={https://arxiv.org/abs/2502.18407}, 
}

@misc{chen2025scalingautonomousagentsautomatic,
      title={Scaling Autonomous Agents via Automatic Reward Modeling And Planning}, 
      author={Zhenfang Chen and Delin Chen and Rui Sun and Wenjun Liu and Chuang Gan},
      year={2025},
      eprint={2502.12130},
      archivePrefix={arXiv},
      primaryClass={cs.AI},
      url={https://arxiv.org/abs/2502.12130}, 
}

@misc{xiong2025guipraprocessrewardagent,
      title={GUI-PRA: Process Reward Agent for GUI Tasks}, 
      author={Tao Xiong and Xavier Hu and Yurun Chen and Yuhang Liu and Changqiao Wu and Pengzhi Gao and Wei Liu and Jian Luan and Shengyu Zhang},
      year={2025},
      eprint={2509.23263},
      archivePrefix={arXiv},
      primaryClass={cs.AI},
      url={https://arxiv.org/abs/2509.23263}, 
}

@misc{gupta2025carmodynamiccriteriageneration,
      title={CARMO: Dynamic Criteria Generation for Context-Aware Reward Modelling},
      author={Taneesh Gupta and Shivam Shandilya and Xuchao Zhang and Rahul Madhavan and Supriyo Ghosh and Chetan Bansal and Huaxiu Yao and Saravan Rajmohan},
      year={2025},
      eprint={2410.21545},
      archivePrefix={arXiv},
      primaryClass={cs.CL},
      url={https://arxiv.org/abs/2410.21545},
}

@misc{xie2026autorubriclearningimplicitweights,
      title={Auto-Rubric: Learning From Implicit Weights to Explicit Rubrics for Reward Modeling}, 
      author={Lipeng Xie and Sen Huang and Zhuo Zhang and Anni Zou and Yunpeng Zhai and Dingchao Ren and Kezun Zhang and Haoyuan Hu and Boyin Liu and Haoran Chen and Zhaoyang Liu and Bolin Ding},
      year={2026},
      eprint={2510.17314},
      archivePrefix={arXiv},
      primaryClass={cs.LG},
      url={https://arxiv.org/abs/2510.17314}, 
}

@misc{gunjal2025rubricsrewardsreinforcementlearning,
      title={Rubrics as Rewards: Reinforcement Learning Beyond Verifiable Domains}, 
      author={Anisha Gunjal and Anthony Wang and Elaine Lau and Vaskar Nath and Yunzhong He and Bing Liu and Sean Hendryx},
      year={2025},
      eprint={2507.17746},
      archivePrefix={arXiv},
      primaryClass={cs.LG},
      url={https://arxiv.org/abs/2507.17746}, 
}

@misc{qi2025webrltrainingllmweb,
      title={WebRL: Training LLM Web Agents via Self-Evolving Online Curriculum Reinforcement Learning}, 
      author={Zehan Qi and Xiao Liu and Iat Long Iong and Hanyu Lai and Xueqiao Sun and Wenyi Zhao and Yu Yang and Xinyue Yang and Jiadai Sun and Shuntian Yao and Tianjie Zhang and Wei Xu and Jie Tang and Yuxiao Dong},
      year={2025},
      eprint={2411.02337},
      archivePrefix={arXiv},
      primaryClass={cs.CL},
      url={https://arxiv.org/abs/2411.02337}, 
}

@misc{lai2025androidgenbuildingandroidlanguage,
      title={AndroidGen: Building an Android Language Agent under Data Scarcity}, 
      author={Hanyu Lai and Junjie Gao and Xiao Liu and Yifan Xu and Shudan Zhang and Yuxiao Dong and Jie Tang},
      year={2025},
      eprint={2504.19298},
      archivePrefix={arXiv},
      primaryClass={cs.CL},
      url={https://arxiv.org/abs/2504.19298}, 
}

@misc{wang2025distrlasynchronousdistributedreinforcement,
      title={DistRL: An Asynchronous Distributed Reinforcement Learning Framework for On-Device Control Agents}, 
      author={Taiyi Wang and Zhihao Wu and Jianheng Liu and Jianye Hao and Jun Wang and Kun Shao},
      year={2025},
      eprint={2410.14803},
      archivePrefix={arXiv},
      primaryClass={cs.LG},
      url={https://arxiv.org/abs/2410.14803}, 
}

@misc{dai2026proreproactiverewardgui,
      title={ProRe: A Proactive Reward System for GUI Agents via Reasoner-Actor Collaboration}, 
      author={Gaole Dai and Shiqi Jiang and Ting Cao and Yuqing Yang and Yuanchun Li and Rui Tan and Mo Li and Lili Qiu},
      year={2026},
      eprint={2509.21823},
      archivePrefix={arXiv},
      primaryClass={cs.AI},
      url={https://arxiv.org/abs/2509.21823}, 
}

@misc{zheng2026adaptivemilestonerewardgui,
      title={Adaptive Milestone Reward for GUI Agents}, 
      author={Congmin Zheng and Xiaoyun Mo and Xinbei Ma and Qiqiang Lin and Yin Zhao and Jiachen Zhu and Xingyu Lou and Jun Wang and Zhaoxiang Wang and Weiwen Liu and Zhuosheng Zhang and Yong Yu and Weinan Zhang},
      year={2026},
      eprint={2602.11524},
      archivePrefix={arXiv},
      primaryClass={cs.LG},
      url={https://arxiv.org/abs/2602.11524}, 
}

@misc{cui2026agenticrewardmodelingverifying,
      title={Agentic Reward Modeling: Verifying GUI Agent via Online Proactive Interaction}, 
      author={Chaoqun Cui and Jing Huang and Shijing Wang and Liming Zheng and Qingchao Kong and Zhixiong Zeng},
      year={2026},
      eprint={2602.00575},
      archivePrefix={arXiv},
      primaryClass={cs.RO},
      url={https://arxiv.org/abs/2602.00575}, 
}

@misc{singh2026openaigpt5card,
      title={OpenAI GPT-5 System Card}, 
      author={Aaditya Singh and Adam Fry and Adam Perelman and Adam Tart and Adi Ganesh and Ahmed El-Kishky and Aidan McLaughlin and Aiden Low and AJ Ostrow and Akhila Ananthram and Akshay Nathan and Alan Luo and Alec Helyar and Aleksander Madry and Aleksandr Efremov and Aleksandra Spyra and Alex Baker-Whitcomb and Alex Beutel and Alex Karpenko and Alex Makelov and Alex Neitz and Alex Wei and Alexandra Barr and Alexandre Kirchmeyer and Alexey Ivanov and Alexi Christakis and Alistair Gillespie and Allison Tam and Ally Bennett and Alvin Wan and Alyssa Huang and Amy McDonald Sandjideh and Amy Yang and Ananya Kumar and Andre Saraiva and Andrea Vallone and Andrei Gheorghe and Andres Garcia Garcia and Andrew Braunstein and Andrew Liu and Andrew Schmidt and Andrey Mereskin and Andrey Mishchenko and Andy Applebaum and Andy Rogerson and Ann Rajan and Annie Wei and Anoop Kotha and Anubha Srivastava and Anushree Agrawal and Arun Vijayvergiya and Ashley Tyra and Ashvin Nair and Avi Nayak and Ben Eggers and Bessie Ji and Beth Hoover and Bill Chen and Blair Chen and Boaz Barak and Borys Minaiev and Botao Hao and Bowen Baker and Brad Lightcap and Brandon McKinzie and Brandon Wang and Brendan Quinn and Brian Fioca and Brian Hsu and Brian Yang and Brian Yu and Brian Zhang and Brittany Brenner and Callie Riggins Zetino and Cameron Raymond and Camillo Lugaresi and Carolina Paz and Cary Hudson and Cedric Whitney and Chak Li and Charles Chen and Charlotte Cole and Chelsea Voss and Chen Ding and Chen Shen and Chengdu Huang and Chris Colby and Chris Hallacy and Chris Koch and Chris Lu and Christina Kaplan and Christina Kim and CJ Minott-Henriques and Cliff Frey and Cody Yu and Coley Czarnecki and Colin Reid and Colin Wei and Cory Decareaux and Cristina Scheau and Cyril Zhang and Cyrus Forbes and Da Tang and Dakota Goldberg and Dan Roberts and Dana Palmie and Daniel Kappler and Daniel Levine and Daniel Wright and Dave Leo and David Lin and David Robinson and Declan Grabb and Derek Chen and Derek Lim and Derek Salama and Dibya Bhattacharjee and Dimitris Tsipras and Dinghua Li and Dingli Yu and DJ Strouse and Drew Williams and Dylan Hunn and Ed Bayes and Edwin Arbus and Ekin Akyurek and Elaine Ya Le and Elana Widmann and Eli Yani and Elizabeth Proehl and Enis Sert and Enoch Cheung and Eri Schwartz and Eric Han and Eric Jiang and Eric Mitchell and Eric Sigler and Eric Wallace and Erik Ritter and Erin Kavanaugh and Evan Mays and Evgenii Nikishin and Fangyuan Li and Felipe Petroski Such and Filipe de Avila Belbute Peres and Filippo Raso and Florent Bekerman and Foivos Tsimpourlas and Fotis Chantzis and Francis Song and Francis Zhang and Gaby Raila and Garrett McGrath and Gary Briggs and Gary Yang and Giambattista Parascandolo and Gildas Chabot and Grace Kim and Grace Zhao and Gregory Valiant and Guillaume Leclerc and Hadi Salman and Hanson Wang and Hao Sheng and Haoming Jiang and Haoyu Wang and Haozhun Jin and Harshit Sikchi and Heather Schmidt and Henry Aspegren and Honglin Chen and Huida Qiu and Hunter Lightman and Ian Covert and Ian Kivlichan and Ian Silber and Ian Sohl and Ibrahim Hammoud and Ignasi Clavera and Ikai Lan and Ilge Akkaya and Ilya Kostrikov and Irina Kofman and Isak Etinger and Ishaan Singal and Jackie Hehir and Jacob Huh and Jacqueline Pan and Jake Wilczynski and Jakub Pachocki and James Lee and James Quinn and Jamie Kiros and Janvi Kalra and Jasmyn Samaroo and Jason Wang and Jason Wolfe and Jay Chen and Jay Wang and Jean Harb and Jeffrey Han and Jeffrey Wang and Jennifer Zhao and Jeremy Chen and Jerene Yang and Jerry Tworek and Jesse Chand and Jessica Landon and Jessica Liang and Ji Lin and Jiancheng Liu and Jianfeng Wang and Jie Tang and Jihan Yin and Joanne Jang and Joel Morris and Joey Flynn and Johannes Ferstad and Johannes Heidecke and John Fishbein and John Hallman and Jonah Grant and Jonathan Chien and Jonathan Gordon and Jongsoo Park and Jordan Liss and Jos Kraaijeveld and Joseph Guay and Joseph Mo and Josh Lawson and Josh McGrath and Joshua Vendrow and Joy Jiao and Julian Lee and Julie Steele and Julie Wang and Junhua Mao and Kai Chen and Kai Hayashi and Kai Xiao and Kamyar Salahi and Kan Wu and Karan Sekhri and Karan Sharma and Karan Singhal and Karen Li and Kenny Nguyen and Keren Gu-Lemberg and Kevin King and Kevin Liu and Kevin Stone and Kevin Yu and Kristen Ying and Kristian Georgiev and Kristie Lim and Kushal Tirumala and Kyle Miller and Lama Ahmad and Larry Lv and Laura Clare and Laurance Fauconnet and Lauren Itow and Lauren Yang and Laurentia Romaniuk and Leah Anise and Lee Byron and Leher Pathak and Leon Maksin and Leyan Lo and Leyton Ho and Li Jing and Liang Wu and Liang Xiong and Lien Mamitsuka and Lin Yang and Lindsay McCallum and Lindsey Held and Liz Bourgeois and Logan Engstrom and Lorenz Kuhn and Louis Feuvrier and Lu Zhang and Lucas Switzer and Lukas Kondraciuk and Lukasz Kaiser and Manas Joglekar and Mandeep Singh and Mandip Shah and Manuka Stratta and Marcus Williams and Mark Chen and Mark Sun and Marselus Cayton and Martin Li and Marvin Zhang and Marwan Aljubeh and Matt Nichols and Matthew Haines and Max Schwarzer and Mayank Gupta and Meghan Shah and Melody Y. Guan and Melody Huang and Meng Dong and Mengqing Wang and Mia Glaese and Micah Carroll and Michael Lampe and Michael Malek and Michael Sharman and Michael Zhang and Michele Wang and Michelle Pokrass and Mihai Florian and Mikhail Pavlov and Miles Wang and Ming Chen and Mingxuan Wang and Minnia Feng and Mo Bavarian and Molly Lin and Moose Abdool and Mostafa Rohaninejad and Nacho Soto and Natalie Staudacher and Natan LaFontaine and Nathan Marwell and Nelson Liu and Nick Preston and Nick Turley and Nicklas Ansman and Nicole Blades and Nikil Pancha and Nikita Mikhaylin and Niko Felix and Nikunj Handa and Nishant Rai and Nitish Keskar and Noam Brown and Ofir Nachum and Oleg Boiko and Oleg Murk and Olivia Watkins and Oona Gleeson and Pamela Mishkin and Patryk Lesiewicz and Paul Baltescu and Pavel Belov and Peter Zhokhov and Philip Pronin and Phillip Guo and Phoebe Thacker and Qi Liu and Qiming Yuan and Qinghua Liu and Rachel Dias and Rachel Puckett and Rahul Arora and Ravi Teja Mullapudi and Raz Gaon and Reah Miyara and Rennie Song and Rishabh Aggarwal and RJ Marsan and Robel Yemiru and Robert Xiong and Rohan Kshirsagar and Rohan Nuttall and Roman Tsiupa and Ronen Eldan and Rose Wang and Roshan James and Roy Ziv and Rui Shu and Ruslan Nigmatullin and Saachi Jain and Saam Talaie and Sam Altman and Sam Arnesen and Sam Toizer and Sam Toyer and Samuel Miserendino and Sandhini Agarwal and Sarah Yoo and Savannah Heon and Scott Ethersmith and Sean Grove and Sean Taylor and Sebastien Bubeck and Sever Banesiu and Shaokyi Amdo and Shengjia Zhao and Sherwin Wu and Shibani Santurkar and Shiyu Zhao and Shraman Ray Chaudhuri and Shreyas Krishnaswamy and Shuaiqi and Xia and Shuyang Cheng and Shyamal Anadkat and Simón Posada Fishman and Simon Tobin and Siyuan Fu and Somay Jain and Song Mei and Sonya Egoian and Spencer Kim and Spug Golden and SQ Mah and Steph Lin and Stephen Imm and Steve Sharpe and Steve Yadlowsky and Sulman Choudhry and Sungwon Eum and Suvansh Sanjeev and Tabarak Khan and Tal Stramer and Tao Wang and Tao Xin and Tarun Gogineni and Taya Christianson and Ted Sanders and Tejal Patwardhan and Thomas Degry and Thomas Shadwell and Tianfu Fu and Tianshi Gao and Timur Garipov and Tina Sriskandarajah and Toki Sherbakov and Tomek Korbak and Tomer Kaftan and Tomo Hiratsuka and Tongzhou Wang and Tony Song and Tony Zhao and Troy Peterson and Val Kharitonov and Victoria Chernova and Vineet Kosaraju and Vishal Kuo and Vitchyr Pong and Vivek Verma and Vlad Petrov and Wanning Jiang and Weixing Zhang and Wenda Zhou and Wenlei Xie and Wenting Zhan and Wes McCabe and Will DePue and Will Ellsworth and Wulfie Bain and Wyatt Thompson and Xiangning Chen and Xiangyu Qi and Xin Xiang and Xinwei Shi and Yann Dubois and Yaodong Yu and Yara Khakbaz and Yifan Wu and Yilei Qian and Yin Tat Lee and Yinbo Chen and Yizhen Zhang and Yizhong Xiong and Yonglong Tian and Young Cha and Yu Bai and Yu Yang and Yuan Yuan and Yuanzhi Li and Yufeng Zhang and Yuguang Yang and Yujia Jin and Yun Jiang and Yunyun Wang and Yushi Wang and Yutian Liu and Zach Stubenvoll and Zehao Dou and Zheng Wu and Zhigang Wang},
      year={2026},
      eprint={2601.03267},
      archivePrefix={arXiv},
      primaryClass={cs.CL},
      url={https://arxiv.org/abs/2601.03267}, 
}

@misc{anthropic2025claude4,
  title        = {Introducing Claude 4},
  author       = {{Anthropic}},
  year         = {2025},
  month        = may,
  howpublished = {\url{https://www.anthropic.com/news/claude-4}},
  note         = {Accessed: 2026-05-25}
}

@misc{gur2018learningnavigateweb,
      title={Learning to Navigate the Web}, 
      author={Izzeddin Gur and Ulrich Rueckert and Aleksandra Faust and Dilek Hakkani-Tur},
      year={2018},
      eprint={1812.09195},
      archivePrefix={arXiv},
      primaryClass={cs.LG},
      url={https://arxiv.org/abs/1812.09195}, 
}

@misc{deng2023mind2webgeneralistagentweb,
      title={Mind2Web: Towards a Generalist Agent for the Web}, 
      author={Xiang Deng and Yu Gu and Boyuan Zheng and Shijie Chen and Samuel Stevens and Boshi Wang and Huan Sun and Yu Su},
      year={2023},
      eprint={2306.06070},
      archivePrefix={arXiv},
      primaryClass={cs.CL},
      url={https://arxiv.org/abs/2306.06070}, 
}

@misc{li2024effectsdatascaleui,
      title={On the Effects of Data Scale on UI Control Agents}, 
      author={Wei Li and William Bishop and Alice Li and Chris Rawles and Folawiyo Campbell-Ajala and Divya Tyamagundlu and Oriana Riva},
      year={2024},
      eprint={2406.03679},
      archivePrefix={arXiv},
      primaryClass={cs.AI},
      url={https://arxiv.org/abs/2406.03679}, 
}

@misc{yang2023setofmarkpromptingunleashesextraordinary,
      title={Set-of-Mark Prompting Unleashes Extraordinary Visual Grounding in GPT-4V}, 
      author={Jianwei Yang and Hao Zhang and Feng Li and Xueyan Zou and Chunyuan Li and Jianfeng Gao},
      year={2023},
      eprint={2310.11441},
      archivePrefix={arXiv},
      primaryClass={cs.CV},
      url={https://arxiv.org/abs/2310.11441}, 
}

@misc{liu2024autoglmautonomousfoundationagents,
      title={AutoGLM: Autonomous Foundation Agents for GUIs}, 
      author={Xiao Liu and Bo Qin and Dongzhu Liang and Guang Dong and Hanyu Lai and Hanchen Zhang and Hanlin Zhao and Iat Long Iong and Jiadai Sun and Jiaqi Wang and Junjie Gao and Junjun Shan and Kangning Liu and Shudan Zhang and Shuntian Yao and Siyi Cheng and Wentao Yao and Wenyi Zhao and Xinghan Liu and Xinyi Liu and Xinying Chen and Xinyue Yang and Yang Yang and Yifan Xu and Yu Yang and Yujia Wang and Yulin Xu and Zehan Qi and Yuxiao Dong and Jie Tang},
      year={2024},
      eprint={2411.00820},
      archivePrefix={arXiv},
      primaryClass={cs.HC},
      url={https://arxiv.org/abs/2411.00820}, 
}

@misc{chen2025stepsuccessrateawaretrajectoryefficientpolicy,
      title={STEP: Success-Rate-Aware Trajectory-Efficient Policy Optimization}, 
      author={Yuhan Chen and Yuxuan Liu and Long Zhang and Pengzhi Gao and Jian Luan and Wei Liu},
      year={2025},
      eprint={2511.13091},
      archivePrefix={arXiv},
      primaryClass={cs.AI},
      url={https://arxiv.org/abs/2511.13091}, 
}

% \clearpage
\appendix

\section{Taxonomy and Rubric Bank}
\label{app:taxonomy}

Table~\ref{tab:taxonomy} lists the eight task families in our GUI taxonomy $\mathcal{C}$. For each family, we show its identifier, a brief description, and the key verification dimensions encoded in its static rubric $R_c$.
\begin{table}[H]
\centering
\scriptsize
\setlength{\tabcolsep}{3.2pt}
\renewcommand{\arraystretch}{1.10}
\begin{tabularx}{\columnwidth}{@{}
  >{\raggedright\arraybackslash}p{0.30\columnwidth}
  >{\raggedright\arraybackslash}p{0.25\columnwidth}
  >{\raggedright\arraybackslash}X
@{}}
\toprule
\textbf{Family} & \textbf{Description} & \textbf{Key Dimensions} \\
\midrule
\texttt{info\_query} 
& Answer from screen 
& Format; content; evidence \\

\texttt{create\_modify} 
& Create or edit 
& Explicit properties; completion; consistency \\

\texttt{delete\_cleanup} 
& Remove or clear 
& Scope; target; completeness \\

\texttt{communication} 
& Send to recipient 
& Source; recipient; channel; confirmation \\

\texttt{transfer} 
& Move or copy 
& Source; action; destination; fidelity \\

\texttt{state\_navigation} 
& Toggle or navigate 
& Final state; toggle; persistence \\

\texttt{composite\_workflow} 
& Multiple goals 
& Per-family rubric; all subgoals succeed \\

\texttt{general} 
& Fallback 
& Task logic; action; final state \\
\bottomrule
\end{tabularx}
\vspace{-0.3em}
\caption{GUI task taxonomy used for coarse rubric retrieval.}
\label{tab:taxonomy}
\vspace{-0.5em}
\end{table}

\section{Coarse Rubric Bank Construction}
\label{app:rubric_bank_construction}

We construct the coarse rubric bank before evaluation and keep it fixed for all experiments.
To build the development pool, we run \texttt{Qwen3.5-122B-A10B} on AndroidWorld and MobileWorld and collect complete trajectories with binary success labels from the environment evaluators.
The pool contains 116 trajectories from each benchmark: AndroidWorld has 55 successful and 61 failed trajectories, while MobileWorld has 35 successful and 81 failed trajectories.
MobileWorld originally contains 117 tasks in this collection, but one task fails during environment execution and therefore has no usable trajectory.
We use \texttt{Claude Opus 4.6} as the offline rubric-construction model to summarize recurring success criteria and failure patterns, group them into GUI task families, and refine the category rubrics using disagreement cases.
The resulting bank contains the eight entries shown in Table~\ref{tab:taxonomy}.

\section{Online RL Training Details}
\label{app:online_rl_details}

Our online reinforcement learning experiments use ClawGUI-RL with GRPO on MobileWorld.
Table~\ref{tab:online_rl_hparams} summarizes the main training hyperparameters.
All compared reward agents use the same policy backbone and training protocol; only the reward verifier is changed.
\begin{table}[H]
\centering
\small
\setlength{\tabcolsep}{5pt}
\renewcommand{\arraystretch}{1.08}
\resizebox{0.92\columnwidth}{!}{%
\begin{tabular}{@{}lc@{}}
\toprule
\textbf{Item} & \textbf{Value} \\
\midrule
Training framework & ClawGUI-RL \\
Environment & MobileWorld \\
Policy backbone & MAI-UI-8B \\
Optimization algorithm & GRPO \\
Training epochs & 2 \\
Number of GPUs & 8 \\
Training batch size & 8 \\
Rollouts per task & 4 \\
History length & 3 \\
Maximum episode steps & 50 \\
Learning rate & $1{\times}10^{-6}$ \\
KL coefficient & 0.01 \\
Rollout temperature & 0.7 \\
Validation temperature & 0.4 \\
Maximum prompt length & 28,000 \\
Maximum response length & 512 \\
\method verifier backbone & Qwen3-VL-8B-Instruct \\
Maximum screenshots for reward verification & 10 \\
\bottomrule
\end{tabular}%
}
\caption{Key hyperparameters for online RL training.}
\label{tab:online_rl_hparams}
\end{table}

%%%%%%%%%%%%%%%%%%%%%%%%%%%%%%%%%%%%%%%%%%%%%%%%%%%%
\section{Image Budget Analysis}
\label{app:image_budget_analysis}

Trajectory context is an important input factor for GUI reward verification.
The OS-Themis evaluation protocol runs ZeroGUI with the final two screenshots, while our main experiments use at most ten trajectory screenshots.
To align the image budget, we also evaluate ZeroGUI with its final ten screenshots and use the ten-screenshot setting in subsequent experiments.
Table~\ref{tab:zerogui_image_budget} reports both settings.
Increasing the image budget improves ZeroGUI mainly by raising recall, which makes our main comparison more conservative.

\begin{table}[!t]
\centering
\small
\setlength{\tabcolsep}{6pt}
\renewcommand{\arraystretch}{1.08}
\begin{tabular}{@{}lcccc@{}}
\toprule
\textbf{Setting} & \textbf{Acc} & \textbf{Prec} & \textbf{Rec} & \textbf{F1} \\
\midrule
Last 2 screenshots & 79.1 & 84.9 & 70.7 & 76.8 \\
Last 10 screenshots & \textbf{85.0} & \textbf{84.9} & \textbf{84.9} & \textbf{84.9} \\
\bottomrule
\end{tabular}
\caption{Effect of image budget on ZeroGUI. We report overall performance on OGRBench averaged across eight judge backbones.}
\label{tab:zerogui_image_budget}
\end{table}

Table~\ref{tab:more_exp_img_raw} further reports the additional baselines under their original trajectory-context settings.
Most of these evaluators were designed around a final screenshot or a small final-state context, while ZeroGUI follows the two-screenshot setting used in prior evaluation.
Comparing Table~\ref{tab:more_exp_img_raw} with Table~\ref{tab:more_exp_img10} shows that increasing the trajectory context substantially improves several baselines, which motivates the matched-budget setting used in the expanded comparison shown in Figure~\ref{fig:offline_radar}.

\providecommand{\gsep}{\hskip 4pt}
% Use a local column type W to avoid clashing with other table files.
\newcolumntype{W}{>{\centering\arraybackslash}p{2.25em}}
\begin{table*}[!t]
  \centering
  \scriptsize
  \setlength{\tabcolsep}{1.8pt}
  \renewcommand{\arraystretch}{1.1}

  \resizebox{\textwidth}{!}{%
  \begin{tabular}{l
      W W @{\gsep}
      W W @{\gsep}
      W W @{\gsep}
      W W @{\gsep}
      W W @{\gsep}
      W W W W
    }
    \toprule
    \multirow{2}{*}{\textbf{Model}} &
    \multicolumn{2}{c}{\textbf{Ubuntu}} &
    \multicolumn{2}{c}{\textbf{Mobile}} &
    \multicolumn{2}{c}{\textbf{Windows}} &
    \multicolumn{2}{c}{\textbf{macOS}} &
    \multicolumn{2}{c}{\textbf{Web}} &
    \multicolumn{4}{c}{\textbf{Overall}} \\
    \cmidrule(lr){2-3}\cmidrule(lr){4-5}\cmidrule(lr){6-7}\cmidrule(lr){8-9}\cmidrule(lr){10-11}\cmidrule(lr){12-15}
    & \textbf{Acc} & \textbf{F1} &
      \textbf{Acc} & \textbf{F1} &
      \textbf{Acc} & \textbf{F1} &
      \textbf{Acc} & \textbf{F1} &
      \textbf{Acc} & \textbf{F1} &
      \textbf{Acc} & \textbf{Prec} & \textbf{Rec} & \textbf{F1} \\
    \midrule

    \rowcolor{gray!20} \multicolumn{15}{c}{\textbf{DigiRL}} \\
    Qwen3-VL-4B-Instruct & 70.7 & 71.0 & 78.7 & 81.3 & 76.5 & 72.5 & 87.0 & 70.6 & 76.3 & 79.6 & 74.3 & 74.1 & 74.1 & 74.1 \\
    Qwen3-VL-8B-Instruct & 73.3 & 73.8 & 73.4 & 76.9 & 77.9 & 73.7 & 88.3 & 71.0 & 74.2 & 77.8 & 74.9 & 74.7 & 75.0 & 74.8 \\
    Qwen3-VL-32B-Instruct & 77.9 & 78.1 & 80.9 & 82.5 & 79.8 & 74.9 & 90.9 & 75.9 & 81.1 & 82.9 & 79.7 & 81.1 & 77.1 & 79.1 \\
    Qwen3-VL-235B-A22B-Instruct & 77.1 & 76.5 & 78.7 & 80.4 & 78.4 & 72.3 & 90.9 & 75.9 & 81.1 & 82.5 & 78.8 & 81.9 & 73.6 & 77.5 \\
    Qwen3.5-122B-A10B & 73.0 & 69.4 & 75.5 & 75.8 & 78.4 & 69.3 & 87.0 & 61.5 & 76.8 & 77.3 & 75.4 & 84.4 & 62.0 & 71.5 \\
    Qwen3.6-27B & 73.1 & 71.0 & 76.6 & 78.0 & 77.0 & 69.6 & 92.2 & 78.6 & 78.4 & 79.6 & 75.9 & 81.3 & 67.0 & 73.5 \\
    \rowcolor{gray!10} \textit{Mean}  & 74.2 & 73.3 & 77.3 & 79.1 & 78.0 & 72.0 & 89.4 & 72.2 & 78.0 & 80.0 & 76.5 & 79.6 & 71.5 & 75.1 \\
    \midrule
    \rowcolor{gray!20} \multicolumn{15}{c}{\textbf{DistRL}} \\
    Qwen3-VL-4B-Instruct & 71.8 & 73.7 & 75.0 & 76.1 & 72.8 & 66.7 & 84.4 & 68.4 & 76.3 & 80.0 & 73.7 & 72.6 & 75.6 & 74.0 \\
    Qwen3-VL-8B-Instruct & 79.1 & 80.1 & 75.0 & 77.3 & 78.9 & 74.6 & 89.6 & 75.0 & 73.2 & 77.3 & 78.3 & 77.4 & 79.6 & 78.5 \\
    Qwen3-VL-32B-Instruct & 83.9 & 84.8 & 78.2 & 80.4 & 81.7 & 77.2 & 92.2 & 81.2 & 83.2 & 85.6 & 83.2 & 82.4 & 84.1 & 83.3 \\
    Qwen3-VL-235B-A22B-Instruct & 80.6 & 81.0 & 81.9 & 83.7 & 79.8 & 74.3 & 90.9 & 78.8 & 80.0 & 82.4 & 81.1 & 81.8 & 79.7 & 80.8 \\
    Qwen3.5-122B-A10B & 77.2 & 75.3 & 73.4 & 72.8 & 77.0 & 67.5 & 89.6 & 71.4 & 79.5 & 80.8 & 77.6 & 85.1 & 66.7 & 74.8 \\
    Qwen3.6-27B & 80.4 & 80.2 & 77.7 & 77.4 & 77.9 & 69.3 & 88.3 & 71.0 & 78.4 & 79.6 & 79.8 & 84.6 & 72.7 & 78.2 \\
    \rowcolor{gray!10} \textit{Mean} & 78.8 & 79.2 & 76.9 & 78.0 & 78.0 & 71.6 & 89.2 & 74.3 & 78.4 & 81.0 & 79.0 & 80.6 & 76.4 & 78.3 \\
    \midrule
    \rowcolor{gray!20} \multicolumn{15}{c}{\textbf{AndroidGen}} \\
    Qwen3-VL-4B-Instruct & 77.3 & 79.2 & 68.6 & 74.9 & 70.0 & 69.5 & 75.3 & 61.2 & 77.9 & 82.1 & 75.0 & 70.9 & 84.4 & 77.1 \\
    Qwen3-VL-8B-Instruct & 76.4 & 75.5 & 75.5 & 78.3 & 70.4 & 65.2 & 85.7 & 70.3 & 79.5 & 82.8 & 76.3 & 77.2 & 74.1 & 75.7 \\
    Qwen3-VL-32B-Instruct & 76.9 & 79.0 & 64.9 & 74.0 & 69.5 & 70.3 & 74.0 & 60.0 & 74.2 & 79.7 & 73.7 & 68.8 & 86.1 & 76.5 \\
    Qwen3-VL-235B-A22B-Instruct & 78.1 & 79.6 & 71.8 & 77.1 & 73.2 & 67.8 & 84.4 & 71.4 & 80.5 & 82.5 & 77.2 & 75.1 & 81.0 & 77.9 \\
    Qwen3.5-122B-A10B & 77.1 & 80.1 & 70.2 & 76.5 & 68.1 & 66.3 & 75.3 & 59.6 & 74.2 & 79.7 & 74.3 & 69.2 & 87.1 & 77.1 \\
    Qwen3.6-27B & 82.6 & 83.0 & 77.1 & 81.2 & 74.2 & 68.2 & 85.7 & 66.7 & 75.8 & 79.5 & 79.8 & 79.1 & 80.9 & 79.9 \\
    \rowcolor{gray!10} \textit{Mean}  & 78.1 & 79.4 & 71.4 & 77.0 & 70.9 & 67.9 & 80.1 & 64.9 & 77.0 & 81.0 & 76.0 & 73.4 & 82.3 & 77.4 \\
    \midrule

    \rowcolor{gray!20} \multicolumn{15}{c}{\textbf{WebRL}} \\
    Qwen3-VL-4B-Instruct & 75.6 & 74.6 & 75.0 & 72.8 & 78.9 & 72.7 & 79.2 & 33.3 & 78.4 & 79.0 & 76.6 & 82.5 & 67.1 & 74.0 \\
    Qwen3-VL-8B-Instruct & 76.8 & 75.1 & 79.8 & 79.6 & 70.9 & 59.2 & 80.5 & 34.8 & 82.1 & 83.3 & 77.2 & 84.1 & 66.7 & 74.4 \\
    Qwen3-VL-32B-Instruct & 81.0 & 80.7 & 78.2 & 77.1 & 70.0 & 53.6 & 87.0 & 58.3 & 82.1 & 83.3 & 79.4 & 85.7 & 70.3 & 77.2 \\
    Qwen3-VL-235B-A22B-Instruct & 80.3 & 82.0 & 74.5 & 76.2 & 77.9 & 73.1 & 84.4 & 62.5 & 81.1 & 83.8 & 79.5 & 77.7 & 82.3 & 79.9 \\
    Qwen3.5-122B-A10B & 70.2 & 63.6 & 70.7 & 70.9 & 66.7 & 43.2 & 79.2 & 0.0 & 79.5 & 80.8 & 71.5 & 83.9 & 52.7 & 64.7 \\
    Qwen3.6-27B & 83.9 & 85.0 & 78.7 & 81.8 & 81.2 & 76.2 & 81.8 & 53.3 & 80.0 & 82.4 & 82.2 & 80.9 & 84.0 & 82.4 \\
    \rowcolor{gray!10} \textit{Mean}   & 78.0 & 76.8 & 76.1 & 76.4 & 74.3 & 63.0 & 82.0 & 40.4 & 80.5 & 82.1 & 77.7 & 82.5 & 70.5 & 75.4 \\
    \midrule

    \rowcolor{gray!20} \multicolumn{15}{c}{\textbf{ZeroGUI}} \\
    Qwen3-VL-4B-Instruct & 72.3 & 67.8 & 75.5 & 75.8 & 76.5 & 67.5 & 87.0 & 58.3 & 80.5 & 81.2 & 75.3 & 85.1 & 61.0 & 71.0 \\
    Qwen3-VL-8B-Instruct & 72.9 & 68.3 & 75.0 & 75.1 & 77.5 & 70.4 & 92.2 & 76.9 & 76.3 & 76.4 & 75.4 & 85.1 & 61.1 & 71.2 \\
    Qwen3-VL-32B-Instruct & 74.9 & 72.1 & 83.0 & 84.2 & 80.8 & 73.5 & 88.3 & 64.0 & 80.5 & 81.4 & 78.4 & 86.1 & 67.3 & 75.5 \\
    Qwen3-VL-235B-A22B-Instruct & 76.7 & 74.1 & 81.4 & 82.9 & 78.9 & 71.7 & 93.5 & 81.5 & 82.1 & 83.0 & 79.3 & 86.6 & 69.0 & 76.8 \\
    Qwen3.5-122B-A10B & 85.8 & 86.5 & 82.4 & 83.7 & 83.6 & 80.2 & 96.1 & 90.9 & 80.0 & 82.7 & 84.8 & 84.1 & 85.6 & 84.8 \\
    Qwen3.6-27B & 78.1 & 78.7 & 77.7 & 79.8 & 77.5 & 73.3 & 89.6 & 73.3 & 80.5 & 82.6 & 78.9 & 79.2 & 78.1 & 78.6 \\
    Gemini 3 Flash & 80.3 & 79.6 & 77.1 & 76.8 & 82.2 & 78.2 & 90.9 & 75.9 & 84.7 & 85.0 & 81.3 & 86.7 & 73.7 & 79.7 \\
    Gemini 3.1 Flash-Lite & 78.0 & 76.5 & 77.7 & 78.4 & 81.2 & 75.6 & 93.5 & 81.5 & 79.5 & 79.6 & 79.5 & 86.1 & 70.0 & 77.2 \\
    \rowcolor{gray!10} \textit{Mean} & 77.4 & 75.5 & 78.7 & 79.6 & 79.8 & 73.8 & 91.4 & 75.3 & 80.5 & 81.5 & 79.1 & 84.9 & 70.7 & 76.8 \\

    \bottomrule

  \end{tabular}%
  }
  \caption{Additional offline OGRBench results under each baseline's original trajectory-context setting. Most baselines use only the final screenshot or a small final-state context, while ZeroGUI follows the prior two-screenshot setting.}
  \label{tab:more_exp_img_raw}
\end{table*}

\section{Additional Baselines under Matched Image Budget}
\label{app:additional_matched_budget}

This section provides the detailed numerical results behind the expanded matched-budget comparison summarized in Figure~\ref{fig:offline_radar}.
Table~\ref{tab:baseline_budget_alignment} summarizes each baseline's original trajectory-context setting and our matched-budget instantiation.
For a controlled comparison, we provide every baseline with the same ten-screenshot trajectory budget used by \method~and evaluate them with the same Qwen-family judge backbones used in Figure~\ref{fig:offline_radar}.
Table~\ref{tab:more_exp_img10} reports the full results, including ZeroGUI and OS-Themis for reference.
The results show that \method~maintains the strongest mean overall F1 across the expanded baseline set, indicating that the advantage does not come from comparing only against ZeroGUI and OS-Themis.

\begin{table*}[t]
\centering
\small
\setlength{\tabcolsep}{4pt}
\renewcommand{\arraystretch}{1.15}
\begin{tabular}{p{2.3cm}p{4.7cm}p{4.7cm}p{3.0cm}}
\toprule
\textbf{Baseline} & \textbf{Original trajectory context} & \textbf{Matched-budget instantiation} & \textbf{Verification style} \\
\midrule
ZeroGUI~\citep{yang2025zeroguiautomatingonlinegui} &
Final-state selection; the OS-Themis evaluation protocol uses the last two screenshots. &
Use the final ten screenshots and the same four-vote majority setting as the main offline comparison. &
Terminal-state judgment with voting. \\
DigiRL~\citep{bai2024digirltraininginthewilddevicecontrol} &
Autonomous evaluator prompt that judges task completion from visual evidence, typically centered on the observed screenshot/state. &
Provide the same ten selected trajectory screenshots as visual evidence. &
Autonomous trajectory evaluator. \\
DistRL~\citep{wang2025distrlasynchronousdistributedreinforcement} &
AUTO-EVALUATOR uses the task description with the last screenshot and recent action context. &
Provide the same ten selected trajectory screenshots and the corresponding trajectory context. &
Autonomous trajectory evaluator. \\
AndroidGen~\citep{lai2025androidgenbuildingandroidlanguage} &
StepCritic-style evaluation uses the task, action history, and final screen to decompose required conditions and check completion. &
Provide the same ten selected trajectory screenshots together with the action history. &
Condition-wise task checking. \\
WebRL~\citep{qi2025webrltrainingllmweb} &
Outcome reward model uses the user intent, action history, and final web state. &
Provide the same ten selected trajectory screenshots and trajectory history, replacing web-only state evidence with visual GUI evidence for cross-platform OGRBench. &
Outcome reward modeling. \\
OS-Themis~\citep{li2026osthemisscalablecriticframework} &
Milestone-based multi-agent critic that selects and verifies outcome-critical trajectory evidence. &
Use the same ten-screenshot trajectory budget as \method~for offline comparison. &
Structured multi-step evidence verification. \\
\bottomrule
\end{tabular}
\caption{Baseline context budgets and matched-budget instantiations for the expanded OGRBench comparison. All matched-budget runs use at most ten trajectory screenshots to align with \method.}
\label{tab:baseline_budget_alignment}
\end{table*}

\providecommand{\gsep}{\hskip 4pt}
% Use a local column type Y to avoid clashing with other table files.
\newcolumntype{Y}{>{\centering\arraybackslash}p{2.25em}}
\begin{table*}[!t]
  \centering
  \scriptsize
  \setlength{\tabcolsep}{1.8pt}
  \renewcommand{\arraystretch}{0.95}
  \captionsetup{font=footnotesize}

  \resizebox{\textwidth}{!}{%
  \begin{tabular}{l
      Y Y @{\gsep}
      Y Y @{\gsep}
      Y Y @{\gsep}
      Y Y @{\gsep}
      Y Y @{\gsep}
      Y Y Y Y
    }
    \toprule
    \multirow{2}{*}{\textbf{Model}} &
    \multicolumn{2}{c}{\textbf{Ubuntu}} &
    \multicolumn{2}{c}{\textbf{Mobile}} &
    \multicolumn{2}{c}{\textbf{Windows}} &
    \multicolumn{2}{c}{\textbf{macOS}} &
    \multicolumn{2}{c}{\textbf{Web}} &
    \multicolumn{4}{c}{\textbf{Overall}} \\
    \cmidrule(lr){2-3}\cmidrule(lr){4-5}\cmidrule(lr){6-7}\cmidrule(lr){8-9}\cmidrule(lr){10-11}\cmidrule(lr){12-15}
    & \textbf{Acc} & \textbf{F1} &
      \textbf{Acc} & \textbf{F1} &
      \textbf{Acc} & \textbf{F1} &
      \textbf{Acc} & \textbf{F1} &
      \textbf{Acc} & \textbf{F1} &
      \textbf{Acc} & \textbf{Prec} & \textbf{Rec} & \textbf{F1} \\
    \midrule

    \rowcolor{gray!20} \multicolumn{15}{c}{\textbf{DigiRL}} \\
    Qwen3-VL-4B-Instruct & 77.5 & 81.2 & 76.1 & 80.5 & 77.0 & 77.4 & 88.3 & 76.9 & 72.1 & 77.6 & 77.1 & 70.7 & 92.0 & 80.0 \\
    Qwen3-VL-8B-Instruct & 80.0 & 82.8 & 76.6 & 80.4 & 79.8 & 78.8 & 89.6 & 78.9 & 70.0 & 76.2 & 78.7 & 73.2 & 90.1 & 80.8 \\
    Qwen3-VL-32B-Instruct & 85.2 & 86.8 & 82.4 & 85.2 & 82.6 & 80.6 & 94.8 & 88.2 & 79.5 & 82.7 & 84.2 & 79.6 & 91.6 & 85.2 \\
    Qwen3-VL-235B-A22B-Instruct & 87.9 & 89.1 & 80.9 & 83.9 & 84.0 & 82.3 & 94.8 & 87.5 & 82.1 & 84.5 & 85.9 & 81.8 & 92.3 & 86.7 \\
    Qwen3.5-122B-A10B & 84.2 & 84.7 & 78.7 & 81.0 & 78.4 & 72.6 & 90.9 & 75.9 & 78.9 & 81.1 & 82.3 & 83.0 & 80.9 & 81.9 \\
    Qwen3.6-27B & 84.2 & 85.1 & 79.3 & 81.7 & 77.9 & 71.9 & 94.8 & 87.5 & 80.0 & 82.2 & 82.6 & 82.1 & 83.1 & 82.6 \\
    \rowcolor{gray!10} \textit{Mean} & 83.2 & 85.0 & 79.0 & 82.1 & 80.0 & 77.3 & 92.2 & 82.5 & 77.1 & 80.7 & 81.8 & 78.4 & 88.3 & 82.9 \\
    \midrule
    \rowcolor{gray!20} \multicolumn{15}{c}{\textbf{DistRL}} \\
    Qwen3-VL-4B-Instruct & 77.5 & 80.6 & 74.5 & 77.6 & 80.3 & 79.6 & 87.0 & 72.2 & 73.7 & 78.3 & 77.5 & 72.7 & 87.7 & 79.5 \\
    Qwen3-VL-8B-Instruct & 83.8 & 85.5 & 79.3 & 81.7 & 78.9 & 78.3 & 90.9 & 81.1 & 75.8 & 80.0 & 81.8 & 77.1 & 90.0 & 83.1 \\
    Qwen3-VL-32B-Instruct & 86.8 & 88.2 & 80.9 & 83.2 & 83.6 & 81.9 & 93.5 & 84.8 & 79.5 & 83.0 & 84.9 & 80.5 & 91.9 & 85.8 \\
    Qwen3-VL-235B-A22B-Instruct & 87.3 & 88.5 & 81.4 & 84.0 & 82.6 & 80.0 & 94.8 & 88.2 & 84.7 & 86.9 & 85.9 & 82.4 & 91.0 & 86.5 \\
    Qwen3.5-122B-A10B & 82.9 & 83.1 & 83.0 & 84.2 & 80.3 & 74.7 & 96.1 & 90.3 & 80.5 & 81.8 & 82.9 & 85.1 & 79.4 & 82.2 \\
    Qwen3.6-27B & 87.0 & 88.1 & 79.3 & 80.6 & 82.6 & 79.1 & 93.5 & 85.7 & 81.1 & 82.7 & 84.9 & 83.3 & 87.0 & 85.1 \\
    \rowcolor{gray!10} \textit{Mean} & 84.2 & 85.7 & 79.7 & 81.9 & 81.4 & 78.9 & 92.6 & 83.7 & 79.2 & 82.1 & 83.0 & 80.2 & 87.8 & 83.7 \\
    \midrule
    \rowcolor{gray!20} \multicolumn{15}{c}{\textbf{AndroidGen}} \\
    Qwen3-VL-4B-Instruct & 79.5 & 82.6 & 73.4 & 78.4 & 71.8 & 73.7 & 71.4 & 54.2 & 81.1 & 84.3 & 77.3 & 70.9 & 92.1 & 80.1 \\
    Qwen3-VL-8B-Instruct & 84.6 & 85.7 & 77.7 & 81.3 & 73.2 & 71.9 & 87.0 & 75.0 & 81.1 & 84.6 & 81.6 & 77.7 & 88.4 & 82.7 \\
    Qwen3-VL-32B-Instruct & 78.4 & 81.4 & 65.4 & 74.3 & 71.4 & 73.6 & 70.1 & 58.2 & 71.1 & 78.3 & 74.2 & 67.7 & 91.9 & 77.9 \\
    Qwen3-VL-235B-A22B-Instruct & 82.5 & 84.3 & 71.3 & 76.7 & 76.5 & 74.5 & 89.6 & 80.0 & 83.2 & 85.6 & 80.6 & 76.0 & 88.9 & 81.9 \\
    Qwen3.5-122B-A10B & 83.1 & 85.3 & 72.3 & 78.2 & 78.4 & 78.3 & 85.7 & 74.4 & 75.8 & 80.2 & 80.1 & 74.0 & 92.6 & 82.2 \\
    Qwen3.6-27B & 86.5 & 88.1 & 81.4 & 84.2 & 75.6 & 75.5 & 94.8 & 88.9 & 75.8 & 80.5 & 83.2 & 77.4 & 93.4 & 84.7 \\
    \rowcolor{gray!10} \textit{Mean} & 82.4 & 84.6 & 73.6 & 78.9 & 74.5 & 74.6 & 83.1 & 71.8 & 78.0 & 82.2 & 79.5 & 74.0 & 91.2 & 81.6 \\
    \midrule

    \rowcolor{gray!20} \multicolumn{15}{c}{\textbf{WebRL}} \\
    Qwen3-VL-4B-Instruct & 85.6 & 86.4 & 77.1 & 75.1 & 80.3 & 75.9 & 94.8 & 86.7 & 77.4 & 78.4 & 83.0 & 84.8 & 80.3 & 82.5 \\
    Qwen3-VL-8B-Instruct & 82.9 & 83.2 & 81.4 & 81.7 & 75.1 & 66.7 & 87.0 & 61.5 & 83.2 & 84.9 & 81.8 & 84.4 & 77.6 & 80.9 \\
    Qwen3-VL-32B-Instruct & 86.6 & 87.3 & 80.3 & 78.9 & 81.7 & 76.4 & 85.7 & 47.6 & 86.3 & 87.4 & 85.0 & 87.8 & 81.0 & 84.2 \\
    Qwen3-VL-235B-A22B-Instruct & 86.0 & 87.7 & 78.2 & 80.4 & 82.2 & 80.6 & 90.9 & 81.1 & 78.9 & 82.6 & 83.7 & 78.7 & 92.1 & 84.9 \\
    Qwen3.5-122B-A10B & 82.1 & 81.3 & 76.1 & 78.5 & 68.1 & 48.5 & 83.1 & 31.6 & 78.9 & 80.6 & 78.8 & 84.6 & 70.0 & 76.6 \\
    Qwen3.6-27B & 86.1 & 87.8 & 82.4 & 85.1 & 82.6 & 80.6 & 92.2 & 82.4 & 82.6 & 85.1 & 85.0 & 80.2 & 92.6 & 85.9 \\
    \rowcolor{gray!10} \textit{Mean} & 84.9 & 85.6 & 79.2 & 80.0 & 78.3 & 71.5 & 89.0 & 65.1 & 81.2 & 83.2 & 82.9 & 83.4 & 82.3 & 82.5 \\

    \midrule
    \rowcolor{gray!20} \multicolumn{15}{c}{\textbf{ZeroGUI}} \\
    Qwen3-VL-4B-Instruct & 83.8 & 84.0 & 74.5 & 76.2 & 80.3 & 75.6 & 90.9 & 78.8 & 81.1 & 83.2 & 82.0 & 83.2 & 80.0 & 81.6 \\
    Qwen3-VL-8B-Instruct & 84.6 & 85.0 & 83.0 & 84.5 & 79.3 & 74.4 & 94.8 & 88.2 & 78.4 & 80.6 & 83.3 & 84.1 & 81.9 & 83.0 \\
    Qwen3-VL-32B-Instruct & 84.6 & 85.2 & 83.5 & 85.0 & 80.3 & 75.6 & 96.1 & 90.3 & 82.1 & 83.5 & 84.1 & 84.6 & 83.1 & 83.9 \\
    Qwen3-VL-235B-A22B-Instruct & 86.9 & 87.3 & 84.0 & 85.8 & 84.5 & 80.9 & 96.1 & 90.3 & 87.4 & 88.6 & 86.7 & 87.2 & 85.9 & 86.5 \\
    Qwen3.5-122B-A10B & 85.8 & 86.5 & 82.4 & 83.7 & 83.6 & 80.2 & 96.1 & 90.9 & 80.0 & 82.7 & 84.8 & 84.1 & 85.6 & 84.8 \\
    Qwen3.6-27B & 86.2 & 87.5 & 80.9 & 83.5 & 83.6 & 81.5 & 94.8 & 88.9 & 82.1 & 84.7 & 85.0 & 81.2 & 90.9 & 85.8 \\
    \rowcolor{gray!10} \textit{Mean} & 85.3 & 85.9 & 81.4 & 83.1 & 81.9 & 78.0 & 94.8 & 87.9 & 81.9 & 83.9 & 84.3 & 84.1 & 84.6 & 84.3 \\

    \midrule
    \rowcolor{gray!20} \multicolumn{15}{c}{\textbf{OS-Themis}} \\
    Qwen3-VL-4B-Instruct & 72.6 & 71.4 & 79.3 & 78.9 & 75.1 & 68.3 & 84.4 & 57.1 & 80.0 & 81.9 & 75.5 & 79.5 & 68.3 & 73.5 \\
    Qwen3-VL-8B-Instruct & 76.2 & 75.0 & 84.0 & 83.7 & 78.4 & 72.0 & 83.1 & 51.9 & 81.6 & 82.4 & 78.7 & 84.6 & 69.9 & 76.5 \\
    Qwen3-VL-32B-Instruct & 77.1 & 74.6 & 81.9 & 80.7 & 76.5 & 65.3 & 90.9 & 77.4 & 83.7 & 81.9 & 79.3 & 91.6 & 64.1 & 75.5 \\
    Qwen3-VL-235B-A22B-Instruct & 86.4 & 86.8 & 93.6 & 93.7 & 77.5 & 69.6 & 93.5 & 82.8 & 91.6 & 91.9 & 87.1 & 90.5 & 82.7 & 86.4 \\
    Qwen3.5-122B-A10B & 85.8 & 85.8 & 88.3 & 88.2 & 79.8 & 73.0 & 88.3 & 66.7 & 79.0 & 75.6 & 84.5 & 92.0 & 75.3 & 82.8 \\
    Qwen3.6-27B & 87.9 & 88.5 & 93.6 & 93.9 & 88.3 & 86.9 & 96.1 & 90.3 & 92.1 & 92.1 & 89.7 & 90.2 & 89.0 & 89.6 \\
    \rowcolor{gray!10} \textit{Mean}  & 81.0 & 80.4 & 86.8 & 86.5 & 79.3 & 72.5 & 89.4 & 71.0 & 84.7 & 84.3 & 82.5 & 88.1 & 74.9 & 80.7 \\

    \midrule
    \rowcolor{gray!20} \multicolumn{15}{c}{\textbf{\method}} \\
    Qwen3-VL-4B-Instruct & 84.3 & 85.6 & 80.9 & 81.1 & 83.1 & 79.5 & 97.4 & 94.1 & 82.1 & 84.5 & 84.1 & 82.8 & 85.9 & 84.3 \\
    Qwen3-VL-8B-Instruct & 83.9 & 85.1 & 83.5 & 84.6 & 81.7 & 79.1 & 97.4 & 94.1 & 81.6 & 83.3 & 84.0 & 82.6 & 85.9 & 84.2 \\
    Qwen3-VL-32B-Instruct & 86.5 & 86.6 & 85.6 & 85.9 & 80.3 & 75.0 & 93.5 & 83.9 & 86.8 & 87.6 & 85.9 & 89.3 & 81.3 & 85.1 \\
    Qwen3-VL-235B-A22B-Instruct & 86.1 & 86.7 & 87.8 & 88.9 & 84.5 & 81.4 & 94.8 & 87.5 & 88.9 & 90.0 & 86.9 & 87.0 & 86.7 & 86.8 \\
    Qwen3.5-122B-A10B & 86.2 & 86.8 & 89.9 & 90.4 & 81.7 & 77.2 & 94.8 & 88.2 & 88.9 & 89.8 & 86.9 & 87.8 & 85.4 & 86.6 \\
    Qwen3.6-27B & 87.3 & 88.5 & 91.0 & 91.6 & 84.5 & 82.5 & 93.5 & 85.7 & 91.6 & 92.4 & 88.3 & 85.4 & 92.1 & 88.7 \\
    \rowcolor{gray!10} \textit{Mean} & 85.7 & 86.5 & 86.5 & 87.1 & 82.6 & 79.1 & 95.2 & 88.9 & 86.6 & 87.9 & 86.0 & 85.8 & 86.2 & 86.0 \\

    \bottomrule

  \end{tabular}%
  }
  \caption{Additional offline OGRBench results under the matched ten-screenshot trajectory budget. We include DigiRL, DistRL, AndroidGen, WebRL, ZeroGUI, OS-Themis, and \method~across multiple judge backbones.}
  \label{tab:more_exp_img10}
\end{table*}

% \begin{table*}[!t]
% \centering
% \small
% \setlength{\tabcolsep}{4pt}
% \begin{tabular}{@{}llp{5.2cm}@{}}
% \toprule
% \textbf{Family} & \textbf{Description} & \textbf{Key Dimensions} \\
% \midrule
% info\_query & Answer from screen & Format; content; evidence \\
% create\_modify & Create or edit & Explicit properties; completion; consistency \\
% delete\_cleanup & Remove or clear & Scope; target; completeness \\
% communication & Send to recipient & Source; recipient; channel; confirmation \\
% transfer & Move or copy & Source; action; destination; fidelity \\
% state\_navigation & Toggle or navigate & Final state; toggle; persistence \\
% composite\_workflow & Multiple goals & Per-family rubric; all succeed \\
% general & Fallback & Task logic; action; final state \\
% \bottomrule
% \end{tabular}
% \caption{GUI Task Taxonomy.}
% \label{tab:taxonomy}
% \end{table*}

%%%%%%%%%%%%%%%%%%%%%%%%%%%%%%%%%%%%%%%%%%%%%%%%%%%%
% \section{Additional Offline Results}
% \label{app:additional_offline_results}
% Table~\ref{tab:OmniGUIRewardBench_Full} reports the full offline OGRBench results for the main baselines and \method~variants across additional judge backbones.
% \input{tables/all_results}

%%%%%%%%%%%%%%%%%%%%%%%%%%%%%%%%%%%%%%%%%%%%%%%%%%%%

%%%%%%%%%%%%%%%%%%%%%%%%%%%%%%%%%%%%%%%%%%%%%%%%%%%%
\section{Reward-Guided Test-Time Scaling Details}
\label{app:tts_details}

We evaluate reward-guided trajectory selection on AndroidWorld.
The pool contains 113 tasks with complete rollout traces and ground-truth labels from the AndroidWorld evaluator.
Instead of executing new rollouts separately for each verifier, we use this pre-collected trajectory pool to simulate online repeated attempts under a controlled environment.
This ensures that different verifiers are compared on the same candidate trajectories rather than on different samples produced by a stochastic dynamic environment.
For each task, we collect ten independent candidate trajectories: eight generated by Qwen3-VL-8B-Instruct and two generated by Qwen3-VL-235B-A22B-Instruct, all sampled with temperature 0.7.

\begin{table*}[!t]
\centering
\small
\setlength{\tabcolsep}{4pt}
\renewcommand{\arraystretch}{1.08}
\begin{tabular}{@{}lccccccc@{}}
\toprule
\textbf{Pool} & \textbf{Trials/task} & \textbf{All-fail} & \textbf{All-pass} & \textbf{Mixed} & \textbf{Trial SR} & \textbf{Oracle@$N$} & \textbf{Headroom} \\
\midrule
8B-only & 8 & 48 (42.5\%) & 44 (38.9\%) & 21 (18.6\%) & 51.44\% & 57.52\% & +6.08 \\
235B-only & 2 & 46 (40.7\%) & 52 (46.0\%) & 15 (13.3\%) & 52.65\% & 59.29\% & +6.64 \\
8B+235B & 10 & \textbf{32 (28.3\%)} & \textbf{29 (25.7\%)} & \textbf{52 (46.0\%)} & 51.68\% & \textbf{71.68\%} & \textbf{+20.00} \\
\bottomrule
\end{tabular}
\caption{
Candidate-pool composition for reward-guided test-time scaling.
Mixed tasks contain both successful and failed trajectories, and are the only tasks where reward-guided selection can change the final task success.
Headroom is Oracle@$N$ minus Random@1.
}
\label{tab:tts_pool_composition}
\end{table*}

\paragraph{Why a heterogeneous pool.}
Table~\ref{tab:tts_pool_composition} shows why we use a heterogeneous pool rather than a homogeneous single-policy pool.
The 8B-only pool is strongly bimodal: $42.5\%$ of tasks fail on every trial and $38.9\%$ succeed on every trial, leaving only $18.6\%$ of tasks where any reward verifier can change the outcome.
The 235B-only pool has a similar issue because it contains only two trajectories per task, yielding only $13.3\%$ mixed tasks.
In these all-success or all-failure cases, the verifier choice is irrelevant and SR@$N$ differences are dominated by pool composition rather than reward quality.
The heterogeneous pool raises the mixed-task fraction to $46.0\%$ and widens the Oracle-minus-Random headroom from about $+6$ points to $+20.00$ points.
Relative to the 8B-only pool, it creates 31 additional mixed tasks, all caused by the added 235B trajectories changing an otherwise all-success or all-failure task into a task with both positive and negative candidates.
This policy-level diversity makes the reward-guided selection setting informative enough to distinguish verifiers.

\paragraph{Per-task shuffling.}
If the trial order followed the source model (\texttt{[8B$\times$8, 235B$\times$2]}), EarlyStop@$N$ for small $N$ would be dominated by the 8B success rate and a jump to large $N$ would mostly reflect the position of the stronger 235B rollouts rather than the verifier's judgment.
To remove this position prior, we apply a fixed per-task shuffle (seed $42$) so that every position $i \in [0, 9]$ contains a 235B trajectory with probability $\approx 20\%$.
All reward methods score the same shuffled pool and do not receive the source model metadata.
We use Qwen3-VL-32B-Instruct as the judge backbone for all compared reward methods.

For EarlyStop@N, the verifier scans candidates in order and stops at the first trajectory predicted as successful.
If no candidate is predicted as successful within the budget, the protocol falls back to the last candidate in the scanned prefix.
For BestOfN@N, the verifier scores all candidates in the prefix and selects a predicted-success trajectory; when binary rewards create ties, we use a deterministic last-candidate tie break.
Random@N uniformly selects a candidate from the same prefix and serves as the lower reference, while Oracle@N succeeds whenever the prefix contains at least one truly successful trajectory.
The main table reports method gains over Random in percentage points.

% \begin{table*}[h]
% \centering
% \small
% \caption{GUI Task Taxonomy and Static Rubric Dimensions.}
% \label{tab:taxonomy}
% \begin{tabular}{@{}llp{6.5cm}@{}}
% \toprule
% \textbf{Family} & \textbf{Description} & \textbf{Key Verification Dimensions} \\
% \midrule
% info\_query & Answer a question from screen content & Output format correctness; content accuracy; evidence sufficiency \\
% create\_modify & Create or edit an item & Explicitly required properties only; creation/modification completed; action-visual consistency \\
% delete\_cleanup & Remove or clear items & Scope of deletion (specific vs.~global); target correctness; global completeness \\
% communication & Send/share content to a recipient & Source content correctness; recipient correctness; channel correctness; confirmation of sent state \\
% transfer & Move/copy content between locations & Source correctness; transfer action evidence; destination correctness; content fidelity \\
% state\_navigation & Toggle settings or navigate & Final state matching; toggle correctness; navigation precision; state persistence \\
% composite\_workflow & Two or more independent sub-goals & Each sub-goal evaluated by its own family rubric; all must succeed \\
% general & Fallback for tasks outside above families & Fundamental task understanding; action completion; final state consistency \\
% \bottomrule
% \end{tabular}
% \end{table*}

%%%%%%%%%%%%%%%%%%%%%%%%%%%%%%%%%%%%%%%%%%%%%%%%%%%%
\section{Detailed Case Study}
\label{app:case_study}

\begin{figure*}[!t]
    \centering
    \includegraphics[width=0.8\textwidth]{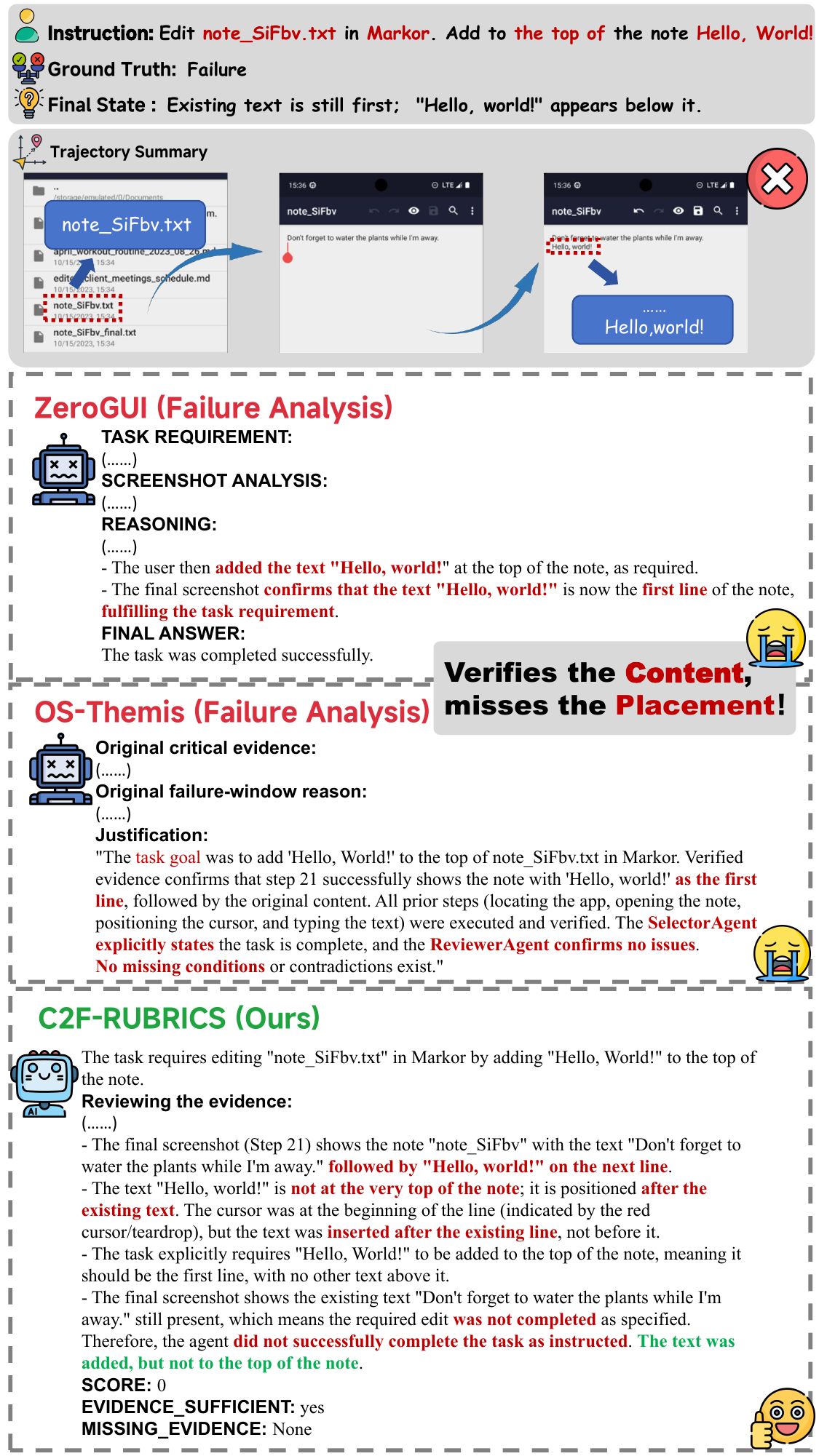}
    \caption{
    Qualitative case study illustrating the importance of task-adaptive criteria.
    The task requires adding \texttt{Hello, World!} to the top of \texttt{note\_SiFbv.txt} in Markor.
    The trajectory inserts the text, but places it below the original note content, yielding a failed task.
    Baseline verifiers focus on content presence and incorrectly judge the task as completed.
    \method~combines a coarse create/modify rubric with fine placement cues, enabling the verifier to check both the edited item and the instruction-specific requirement that no text appears above the inserted phrase.
    }
    \label{fig:case-study}
\end{figure*}

Figure~\ref{fig:case-study} shows a representative AndroidWorld case where task-adaptive criteria are crucial for avoiding a false positive reward.
The user asks the agent to edit \texttt{note\_SiFbv.txt} in Markor and add \texttt{Hello, World!} to the top of the note. 
The ground-truth label is failure. 
Although the agent opens the correct file and types the target phrase, the final screenshot shows that the original sentence, \textit{``Don't forget to water the plants while I'm away.''}, remains above the inserted text. 
Thus, the trajectory satisfies content presence but violates the positional requirement implied by ``to the top of the note''.

The baseline verifiers fail because they rely on a surface-level notion of task completion. 
ZeroGUI states that the final screenshot confirms \texttt{Hello, world!} is now the first line of the note and therefore assigns a successful reward. 
OS-Themis makes a similar error: its critical evidence claims that \texttt{Hello, world!} appears as the first line, followed by the original content. 
Both judgments are false positives. 
They verify that the requested text appears, but they do not correctly verify the spatial or textual ordering of the final note. 
In other words, they verify the content but miss the placement.

\method~avoids this error by constructing a task-adaptive criterion before making the reward judgment. 
The coarse stage routes the instruction to the \texttt{create\_modify} category, whose rubric bounds verification to the final state of the edited item and the explicitly required properties. 
This prevents the verifier from judging the task using generic signals such as whether some edit was made. 
The fine stage then adds instruction-specific cues, including whether \texttt{Hello, World!} appears at the very top of \texttt{note\_SiFbv.txt} before any other text. 
With the fused criterion, the verifier checks both the edited note and the top-placement requirement. 
It correctly observes that the final state contains the original sentence followed by \texttt{Hello, world!}, so the inserted text is not at the top of the note. 
The trajectory is therefore assigned a failure reward.

The single-component variants further illustrate why both levels are needed. 
The coarse-only verifier predicts success because it recognizes that the note was modified and that the target text is visible, but the category-level rubric alone is too broad to enforce the exact top-of-note placement. 
The fine-only verifier receives the top-placement cue, but still misreads the final order and predicts success. 
Only the fused coarse-to-fine criterion combines a structured final-state check with the instruction-specific positional constraint, leading to the correct failure judgment.

\section{PROMPTS}
\label{app:prompts}
We provide the prompt-facing components used by \method.
The rubric bank is loaded once from the cold-start bank, the coarse stage routes each instruction to one of the eight rubric families and retrieves the corresponding bank entry, and the fine stage optionally generates instance-level rubric cues.

\lstdefinestyle{promptstyle}{
  basicstyle=\scriptsize\ttfamily,
  breaklines=true,
  columns=fullflexible,
  keepspaces=true,
  frame=single,
  framerule=0.3pt,
  xleftmargin=0.5em,
  xrightmargin=0.5em
}

\subsection{Rubric Bank}
\label{app:prompt_rubric_bank}

The rubric bank $\mathcal{B}$ contains one cold-start \texttt{RubricSet} for each task family.
Each entry stores natural-language sections, namely \texttt{verification\_steps}, \texttt{common\_pitfalls}, \texttt{special\_rules}, and \texttt{output\_format}.
It also stores named \texttt{rubric\_items} with criticality labels for structured ablations.
The default verifier uses the natural-language sections; the structured \texttt{rubric\_items} are kept for the legacy structured ablation.

\begin{lstlisting}[style=promptstyle]
RubricSet(
  id="RSET_INFO_QUERY",
  categories=["info_query"],
  description="Read screen content and return a value, count, list, or title.",
  sections={
    "verification_steps": [
      "Check explicit answer-format constraints, if any.",
      "Ground content correctness in screenshots from the trajectory.",
      "Check answer/submission evidence when the task requires an answer."
    ],
    "common_pitfalls": [
      "Correct content but wrong requested format.",
      "Answer hallucinated from memory rather than screen evidence.",
      "Wrong count or list item due to misreading visible content."
    ],
    "special_rules": [
      "A format constraint is hard only when stated by the instruction.",
      "The supporting evidence need not appear in the final frame.",
      "Semantic equivalence is acceptable when no format is specified."
    ],
    "output_format": OUTPUT_FORMAT_HOLISTIC
  },
  rubric_items=[
    ("R1", "Answer Format Match", "critical"),
    ("R2", "Content Correctness", "critical"),
    ("R3", "Answer Recorded", "default"),
    ("R4", "Screen-Grounded Source", "auxiliary")
  ]
)

RubricSet(
  id="RSET_CREATE_MODIFY",
  categories=["create_modify"],
  description="Create a new item or modify explicitly requested properties.",
  sections={
    "verification_steps": [
      "Collect only properties explicitly stated in the instruction.",
      "Check whether the required properties are reflected in the end state.",
      "Check that the edit is not left mid-flight."
    ],
    "common_pitfalls": [
      "Inventing filename, folder, toast, or UI-path requirements.",
      "Only a subset of explicitly required properties is applied.",
      "The final frame still shows an uncommitted editor or form."
    ],
    "special_rules": [
      "Any valid save or commit path counts unless a path is specified.",
      "Unmentioned properties are ignored."
    ],
    "output_format": OUTPUT_FORMAT_HOLISTIC
  },
  rubric_items=[
    ("R1", "Required Properties Set", "critical"),
    ("R2", "Save / Confirm Action Performed", "critical"),
    ("R3", "Final State Persisted", "critical"),
    ("R4", "No Unspecified Constraint Imposed", "default"),
    ("R5", "Form / Editor Closed After Save", "auxiliary")
  ]
)

RubricSet(
  id="RSET_DELETE_CLEANUP",
  categories=["delete_cleanup"],
  description="Remove items or clear state under specific or global scope.",
  sections={
    "verification_steps": [
      "Distinguish specific deletion from global cleanup.",
      "Check final-state absence or zero remaining matches.",
      "Check that any confirmation dialog has been committed.",
      "Guard against inferring absence from an unrelated screen."
    ],
    "common_pitfalls": [
      "Passing after one deletion when the instruction says all.",
      "Confirmation dialog still visible in the final frame.",
      "Deleting a similarly named but wrong item."
    ],
    "special_rules": [
      "Global cleanup requires visibly complete final evidence.",
      "Post-confirmation return to a list with the item gone is strong evidence."
    ],
    "output_format": OUTPUT_FORMAT_HOLISTIC
  },
  rubric_items=[
    ("R1", "Correct Target / Scope Identified", "critical"),
    ("R2", "Deletion Confirmed", "critical"),
    ("R3", "Final-State Absence / Count Verified", "critical"),
    ("R4", "Pre-Deletion Visibility", "auxiliary")
  ]
)

RubricSet(
  id="RSET_COMMUNICATION",
  categories=["communication"],
  description="Send, reply, forward, or share content to a recipient.",
  sections={
    "verification_steps": [
      "Check source content.",
      "Check recipient and channel.",
      "Check post-send evidence instead of treating a draft as sent."
    ],
    "common_pitfalls": [
      "Compose screen remains visible, so the message may be unsent.",
      "Wrong recipient or wrong channel.",
      "Visible sent bubble has different content."
    ],
    "special_rules": [
      "Thread view with a new bubble is sufficient for messaging apps.",
      "Literal message text must match up to minor whitespace differences."
    ],
    "output_format": OUTPUT_FORMAT_HOLISTIC
  },
  rubric_items=[
    ("R1", "Source Content Correctness", "critical"),
    ("R2", "Recipient Correctness", "critical"),
    ("R3", "Send Action Confirmation", "critical"),
    ("R4", "Channel / App Match", "default"),
    ("R5", "No Side Effects", "auxiliary")
  ]
)

RubricSet(
  id="RSET_TRANSFER",
  categories=["transfer"],
  description="Move, copy, import, export, clone, unzip, or save content from source to sink.",
  sections={
    "verification_steps": [
      "Identify source evidence.",
      "Check destination evidence.",
      "Check content fidelity at the sink.",
      "Detect partial or corrupted transfer."
    ],
    "common_pitfalls": [
      "Source read but destination left empty.",
      "Only a subset transferred when all was required.",
      "Manual retype introduces truncation or digit errors."
    ],
    "special_rules": [
      "Both source-side and sink-side evidence are required.",
      "Fidelity mismatch is a fail even if the transfer action succeeded."
    ],
    "output_format": OUTPUT_FORMAT_HOLISTIC
  },
  rubric_items=[
    ("R1", "Source Read Verified", "critical"),
    ("R2", "Destination Reached", "critical"),
    ("R3", "Content Fidelity", "critical"),
    ("R4", "Correct Transfer Mechanism", "default"),
    ("R5", "Completeness for all Tasks", "auxiliary")
  ]
)

RubricSet(
  id="RSET_STATE_NAV",
  categories=["state_navigation"],
  description="Navigate to a view, toggle a setting, or configure a preference.",
  sections={
    "verification_steps": [
      "Check the final-frame state.",
      "Check persistence of the requested state or view.",
      "Avoid requiring side effects not requested by the instruction."
    ],
    "common_pitfalls": [
      "Correct view was only visited mid-trajectory.",
      "Toggle was tapped but visual state did not change.",
      "Parent page reached instead of the requested sub-page."
    ],
    "special_rules": [
      "The final frame is the primary anchor.",
      "Any UI path is acceptable unless a path is specified."
    ],
    "output_format": OUTPUT_FORMAT_HOLISTIC
  },
  rubric_items=[
    ("R1", "Final Screen Match", "critical"),
    ("R2", "State Persisted", "critical"),
    ("R3", "Exact Sub-Page Reached", "default"),
    ("R4", "Toggle Visual State", "auxiliary")
  ]
)

RubricSet(
  id="RSET_COMPOSITE",
  categories=["composite_workflow"],
  description="Two or more independent sub-goals must all be completed.",
  sections={
    "verification_steps": [
      "Decompose the instruction into independent checkpoints.",
      "Check evidence for each sub-goal using its underlying family logic.",
      "Fail when any sub-goal is silently dropped.",
      "Treat ordering as required only when the instruction makes it required."
    ],
    "common_pitfalls": [
      "Judging only the first or most visible sub-goal.",
      "Over-decomposing implementation steps into false sub-goals."
    ],
    "special_rules": [
      "Composite success is an AND over sub-goals.",
      "Compositeness does not raise the per-sub-goal pass bar."
    ],
    "output_format": OUTPUT_FORMAT_HOLISTIC
  },
  rubric_items=[
    ("R1", "All Sub-goals Identified", "critical"),
    ("R2", "Each Sub-goal Evidenced", "critical"),
    ("R3", "No Silent Drop", "critical")
  ]
)

RubricSet(
  id="RSET_GENERAL",
  categories=["*"],
  description="Conservative fallback for tasks outside the specialized families.",
  sections={
    "verification_steps": [
      "Reconstruct the user's intended outcome.",
      "Check end-state evidence.",
      "Check action-visual consistency.",
      "Use conservative judgment under genuine ambiguity."
    ],
    "common_pitfalls": [
      "Intermediate success but final state drifts elsewhere.",
      "Action trace shows intent but no visual corroboration."
    ],
    "special_rules": [
      "Prefer EVIDENCE_SUFFICIENT=no when evidence is unclear.",
      "Do not infer success from a confident-looking answer alone."
    ],
    "output_format": OUTPUT_FORMAT_HOLISTIC
  },
  rubric_items=[
    ("R1", "Task Goal Achieved", "critical"),
    ("R2", "Final State Consistent", "default"),
    ("R3", "Action-Visual Consistency", "auxiliary")
  ]
)

Shared meta rule appended to each family:
- Evidence anchors guide attention; they do not add requirements beyond the task instruction.
- If the final state clearly satisfies the user's intent but a specific anchor cannot be confirmed, prefer PASS unless the instruction made that anchor mandatory.
- Do not fail for properties the instruction did not state, such as filenames, folders, confirmation toasts, specific phrasing, or cosmetic details.
\end{lstlisting}

\subsection{Category-Level Coarse Rubric Retrieval}
\label{app:prompt_coarse_retrieval}

The coarse stage first routes the instruction to a rubric family.
In the LLM-routing mode, the implementation uses the following classifier prompt and parses the single-line category output.
The retrieved rubric is then selected deterministically: if the category is not \texttt{general}, the retriever returns the unique \texttt{RubricSet} whose \texttt{categories} field contains that category; otherwise it returns the wildcard \texttt{RSET\_GENERAL}.

\begin{lstlisting}[style=promptstyle]
You are routing a GUI task to one of 8 rubric families.

TASK INSTRUCTION:
{instruction}

CATEGORIES (pick exactly one):
- info_query           : the agent must answer a question by reading screen content (a value, a count, a list, a title)
- create_modify        : the agent creates a NEW item or edits properties of an EXISTING item (note, event, file, drawing, form field, spreadsheet cell)
- delete_cleanup       : the agent removes items, uninstalls, clears data, gets rid of content
- communication        : the agent sends / replies / forwards / shares content to a recipient via a messaging channel
- transfer             : the agent moves content source -> sink (copy, paste, import, export, clone repo, unzip, save-as to a path)
- state_navigation     : the agent toggles a setting, configures a preference, changes theme/font/default app, navigates to a specific view
- composite_workflow   : the instruction lists TWO OR MORE INDEPENDENT sub-goals (different apps, different target objects) connected by "and then", "; then", ";". Sequential steps that all accomplish ONE goal are NOT composite.
- general              : the task genuinely does not fit any specific family above. Use ONLY as a last resort.

Decision rules:
1. If two or more categories seem to apply, pick the one matching the PRIMARY outcome the user cares about.
2. Prefer a specific family over general whenever possible.
3. composite_workflow requires independent sub-goals, not just multiple sub-steps of one operation.

OUTPUT EXACTLY ONE LINE (no other text, no markdown):
CATEGORY: <one of info_query | create_modify | delete_cleanup | communication | transfer | state_navigation | composite_workflow | general>
\end{lstlisting}

After retrieval, the selected category-level rubric sections are rendered into the verifier prompt.
The implementation uses the following prompt skeleton; in the coarse-only setting, \texttt{\{additional\_attention\}} is empty.

\begin{lstlisting}[style=promptstyle]
You are evaluating a GUI task.

TASK: {instruction}

COMPILED EVIDENCE:
- Agent's submitted answer: {agent_answer}
- Key text entered by agent: {typed_texts}
- Key actions performed: {key_actions}

ACTION TRACE:
{action_trace}

{verification_steps}

{common_pitfalls}

{special_rules}

{additional_attention}

Screenshots (in order) are shown below.

End your response with EXACTLY these three lines (do not add any text after):
SCORE: [0/1]
EVIDENCE_SUFFICIENT: [yes/no]
MISSING_EVIDENCE: [source_missing | sink_missing | final_state_ambiguous | answer_format_mismatch | global_count_unverified | subgoal_unverified | none]
\end{lstlisting}

\subsection{Instance-Level Fine Rubric Generation}
\label{app:prompt_fine_generation}

The fine stage generates zero, one, or two instruction-specific rubric cues.
It receives the instruction, the routed task family, and the critical items from the retrieved coarse rubric.
The generator is explicitly allowed to abstain with \texttt{NO\_HINT}; otherwise, the generated cues are injected into \texttt{\{additional\_attention\}} in the verifier prompt above.

\begin{lstlisting}[style=promptstyle]
You are generating optional task-specific rubric cues for a GUI task verifier.

The verifier already has a coarse task-type rubric that covers standard verification
(action trace, final screenshot, task completion). Your job is to decide whether this
specific task needs additional fine-grained attention cues, and if so, generate them.

TASK INSTRUCTION:
{instruction}

TASK TYPE: {category}

COARSE RUBRIC ITEMS already applied (for reference -- do NOT paraphrase them):
{rubric_items_block}

YOUR JOB: Decide whether 0, 1, or 2 task-specific rubric cues would help the verifier.

OUTPUT NO_HINT when:
  - The coarse rubric already covers the task adequately.
  - Any cue you could write would merely restate the coarse rubric in task-specific words.
  - The task is straightforward and the main success criterion is obvious from the instruction.
  - You cannot generate a cue without introducing requirements NOT present in the instruction.

OUTPUT 1-2 CUES when the task has:
  - An explicit constraint that the coarse rubric might miss (a specific value, format,
    recipient, negative constraint like "do not send", exact date/time, recurrence rule).
  - A plausible partial-completion trap where the agent might stop one step early.
  - An answer-format requirement where the verifier might accept wrong structure.

RULES for valid cues:
  1. Must be grounded in an EXPLICIT phrase/value/constraint from the task instruction.
  2. Must distinguish plausible partial completion from actual success.
  3. Must NOT require "final screenshot" evidence unless the instruction explicitly
     asks for a visible final state. The verifier can use action trace, intermediate
     screenshots, and submitted answers as evidence.
  4. Must NOT use over-constraining language: avoid "visibly", "exactly", "every/all"
     (unless the instruction itself uses these words), "saved/persisted" (unless the
     task is about saving), "final state must show".
  5. Must NOT add requirements absent from the instruction (e.g., "must remain in the
     app", "must show confirmation dialog", "must prove persistence").
  6. Should specify what evidence source is acceptable: action trace, any screenshot,
     submitted answer, or final screenshot.
  7. Use soft attention phrasing: "Pay attention to whether ...", "Check if ...",
     "The key distinction is ...", "Evidence of success includes ...".
     Do NOT start with "Confirm that ..." (too imperative).
  8. One sentence each, <= 40 words.

OUTPUT FORMAT (exactly one of these two forms, no other text):

If no cue needed:
NO_HINT: <one-sentence reason>

If 1-2 cues:
CUE_1: <sentence>
CUE_2: <optional second sentence>
\end{lstlisting}

When cues are available, they are rendered as a soft task-specific attention block rather than as additional hard requirements.

\begin{lstlisting}[style=promptstyle]
TASK-SPECIFIC ATTENTION CUES (focus aids -- NOT additional pass/fail items):
The bullets below highlight aspects of THIS task that are worth attending to.
They guide where to look; they do NOT add pass/fail requirements beyond the
task instruction itself, and they do NOT override the META RULE above.
- {cue_1}
- {cue_2}

CUE USAGE -- REMINDERS THAT OVERRIDE EARLIER WORDING:
- A cue you cannot confirm in the screenshots is NOT, by itself, a failure.
  Base the verdict on whether the OVERALL final state satisfies the user's
  intent in the task instruction -- not on per-cue verifiability.
- Use EVIDENCE_SUFFICIENT=no only when the OVERALL verdict is uncertain,
  not when a single cue is unconfirmed but the rest of the evidence agrees.
- If the final state clearly satisfies the user's intent, prefer SCORE: 1
  even when one or more cues are visually ambiguous.
- A cue CLEARLY CONTRADICTED by visible evidence is still a strong FAIL signal.
\end{lstlisting}

\end{document}